\documentclass[lettersize,journal]{IEEEtran}
\usepackage{amsmath,amsfonts}
\usepackage{algorithmic}
\usepackage{array}
\usepackage[caption=false,font=normalsize,labelfont=sf,textfont=sf]{subfig}
\usepackage{textcomp}
\usepackage{stfloats}
\usepackage{url}
\usepackage{verbatim}
\usepackage{graphicx}
\usepackage{cite}
\usepackage{booktabs}       

\usepackage[utf8]{inputenc} 
\usepackage[T1]{fontenc}    
\usepackage{url}            
\usepackage{amsfonts}       
\usepackage{nicefrac}       
\usepackage{microtype}      
\usepackage{xcolor}         
\usepackage{epsfig}
\usepackage{amsmath}
\usepackage{amssymb}
\usepackage{caption}
\usepackage{tabularx}

\usepackage{multirow}
\newcommand{\cmark}{\checkmark}

\usepackage[linesnumberedhidden,ruled]{algorithm2e}

\newcommand{\bx}{\ensuremath{{\boldsymbol x}}}

\newcommand{\by}{\ensuremath{{\boldsymbol y}}}

\newcommand{\mcalM}{\ensuremath{{\mathcal{M}}}}

\newcommand{\mcalF}{\ensuremath{{\mathcal{F}}}}
\newcommand{\mcalQ}{\ensuremath{{\mathcal{Q}}}}

\newcommand{\mcalT}{\ensuremath{{\mathcal{T}}}}

\newcommand{\bz}{\ensuremath{{\boldsymbol z}}}

\begin{document}

\title{Lifelong Person Search}

\author{Jae-Won Yang, Seungbin Hong, and Jae-Young Sim,~\IEEEmembership{\normalsize Member,~IEEE}
}

\maketitle

\begin{abstract}
  Person search is the task to localize a query person in gallery datasets of scene images. Existing methods have been mainly developed to handle a single target dataset only, however diverse datasets are continuously given in practical applications of person search. In such cases, they suffer from the catastrophic knowledge forgetting in the old datasets when trained on new datasets. In this paper, we first introduce a novel problem of lifelong person search (LPS) where the model is incrementally trained on the new datasets while preserving the knowledge learned in the old datasets. We propose an end-to-end LPS framework that facilitates the knowledge distillation to enforce the consistency learning between the old and new models by utilizing the prototype features of the foreground persons as well as the hard background proposals in the old domains. Moreover, we also devise the rehearsal-based instance matching to further improve the discrimination ability in the old domains by using the unlabeled person instances additionally. Experimental results demonstrate that the proposed method achieves significantly superior performance of both the detection and re-identification to preserve the knowledge learned in the old domains compared with the existing methods.
\end{abstract}


\section{Introduction}
Person search is the technique to find the query person from the gallery sets of scene images where multiple persons usually appear simultaneously in each image. It has been drawing much attention due to its practical applicability to various real-world scenarios such as large-scale video understanding, surveillance, and augmented reality.
Different from the person re-identification (re-ID)~\cite{reidfirst},~\cite{li2018harmonious},~\cite{luo2019strong},~\cite{ning2020feature},~\cite{liu2022neural},~\cite{huang2020multiscale},~\cite{chen2022saliency} that finds the query person from the sets of cropped person images, the person search is a more challenging task that first localizes the bounding boxes of person instances in the scene images and then matches the identities of the detected instances to the query person. The person search can be implemented in the two-step manner by using separately trained two sub-networks of the object detection and re-ID. However, the training of the two-step methods is usually inefficient requiring huge computational complexity, since the detection network extracts the bounding boxes for the person instances from the scene images which are then inputted to the re-ID network to retrieve the features again tailored to the re-ID task.

To overcome this issue, the end-to-end learning was introduced that jointly trains the person detection and re-ID networks. The end-to-end methods have been mainly developed in the supervised learning manner based on the assumption that both of the training and test data come from a same target dataset~\cite{norm_aware},~\cite{kim2021prototype},~\cite{yan2021anchor},~\cite{han2021end},~\cite{lee2022oimnet},~\cite{yu2022cascade}. However, in many practical real-world applications, multiple datasets are generated in different places and times that exhibit domain gaps from one another. In such cases, the existing supervised methods trained on a certain dataset usually fail to work on other datasets. Furthermore, re-training the network, whenever the target datasets are changed, suffers from the high computational complexity as well as the catastrophic forgetting~\cite{kirkpatrick2017overcoming} of the knowledge learned from the previously trained datasets.

\begin{figure}[t]
	\centering
	\includegraphics[width=0.99\linewidth]{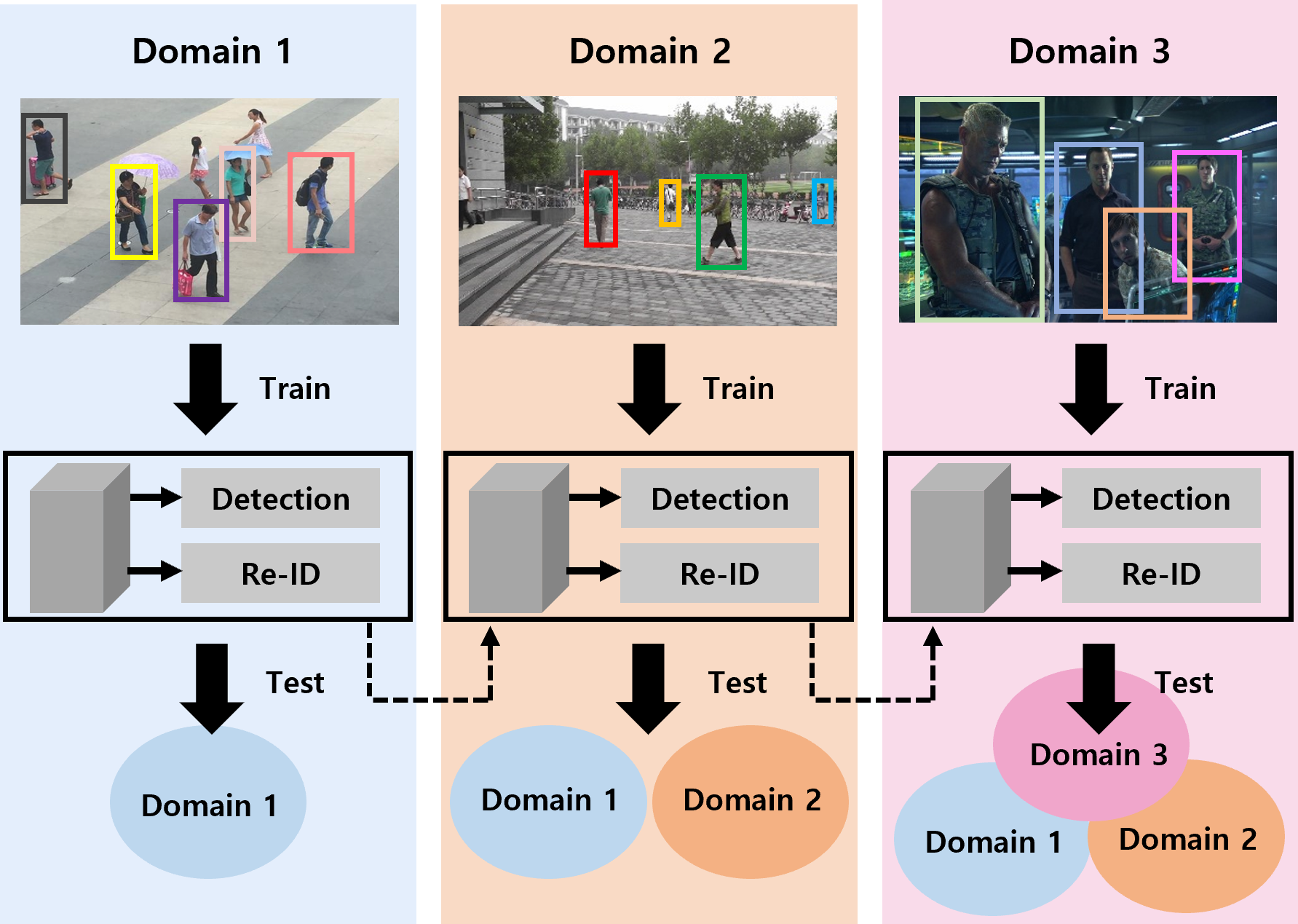}
	\caption{The concept of the proposed lifelong person search. New datasets on different domains are given in order. The model is incrementally trained on new domains without entire re-training on all the datasets while preserving the knowledge of old domains.}
	\label{fig:introduction}
\end{figure}

In this paper, we first introduce a new problem of lifelong person search (LPS) where the new datasets on different domains are assumed to be sequentially given in order, as shown in Fig.~\ref{fig:introduction}. The model is forced to be generalized to all domains while preserving the previously learned knowledge without entire re-training using all the datasets. 
The end-to-end LPS is more challenging compared to the lifelong object detection~\cite{shmelkov2017incremental},~\cite{peng2020faster} and lifelong person re-ID~\cite{wu2021generalising},~\cite{pu2021lifelong},~\cite{ge2022lifelong},~\cite{huang2022lifelong}, since it suffers from the catastrophic forgetting problem in both sub-tasks of the person detection and re-ID.
Whereas the lifelong person detection is a domain-incremental task where only the same class of person is localized across different domains, the lifelong person re-ID is related to both domain-incremental and class-incremental tasks since the new person identities are additionally given from different domains.
Moreover, the end-to-end person search also suffers from the task conflict problem where the person detection focuses on extracting the common representation of persons distinct from the backgrounds, but the person re-ID attempts to extract the unique representations according to the person identities. This task conflict problem becomes more serious in LPS scenario where the model is encouraged to be continuously adapted to different domains. 
In addition, the lifelong re-ID methods~\cite{wu2021generalising},~\cite{pu2021lifelong},~\cite{ ge2022lifelong},~\cite{huang2022lifelong} have been usually developed for full-body pedestrian images on similar domains. 
However, as shown in Fig.~\ref{fig:introduction}, the LPS considers the scene images with severely different characteristics across multiple domains, for example, diverse backgrounds, different scales and densities of persons, and even local body parts due to partial occlusion. 

To address the LPS problem, we propose a novel end-to-end framework that generalizes the network to be incrementally adapted to the new domains while preserving the previously learned knowledge in the old domains.
Specifically, we perform the knowledge distillation for both sub-tasks of the person detection and re-ID using the old exemplar data. We first use the prototypes, representative features of person identities, associated with the old exemplar data, to design the rehearsal-based re-ID knowledge distillation loss that enforces the consistency on the distributions of the feature similarity between the old and new models. 
Moreover, we also utilize the hard background proposals additionally to refine the re-ID knowledge distillation loss that alleviate the effect of inaccurately detected person proposals and extract discriminative features for re-ID more reliably.   
In addition, we devise the rehearsal-based instance matching loss to further improve the model's discrimination ability. We minimize the feature discrepancy between the labeled proposals in the old exemplar data and its ground truth old prototypes. We also employ the unlabeled person identities in the old domains as negative samples to preserve the knowledge effectively. Experimental results demonstrate that the proposed method preserves the knowledge of old datasets more faithfully compared with the existing methods, and therefore serves as a very promising tool for LPS.   

The main contributions of this paper are summarized as follows.
\begin{itemize}
	\item To the best of our knowledge, we first introduce a new and challenging problem of person search with lifelong learning scenario where the model is incrementally trained on the new domains while preserving the previously learned knowledge in the old domains. 
	\item We propose an end-to-end LPS framework that jointly trains the detection and re-ID networks by using the rehearsal-based knowledge distillation loss and the instance matching loss, that alleviate the catastrophic forgetting of the old knowledge during the training on the new domains.
	\item We demonstrate the efficiency of the proposed method by providing comprehensive experimental results compared with the existing methods based on the lifelong learning scenario.
\end{itemize}

\section{Related Works}
\subsection{Person Search}
Person search has been studied mainly in the supervised manner where the bounding boxes of person instances and the person identities are labeled in the training datasets.
The two-step methods train the person detection and re-ID networks separately to prevent the conflict problem between the two tasks. Zheng \textit{et al.}~\cite{prw} conducted extensive experiments by training the state-of-the-art methods of the pedestrian detection and person re-ID. They also provided a benchmark PRW dataset. Pu~\textit{et al.}~\cite{chen2018person} proposed a segmentation masking scheme to force the re-ID network to focus on the foreground regions of the detected persons. Lan~\textit{et al.}~\cite{lan2018eccv} extracted multi-scale features to deal with the scale variation problem of person size in the scene images. Wang~\textit{et al.}~\cite{wang2020tcts} made the re-ID network more adapted to the detection results by composing the training set with the person images cropped by the pre-trained detection network and the person images cropped by using the bounding box labels. Ke~\textit{et al.}~\cite{ke2022joint} performed a data augmentation scheme that shifts the locations of the ground truth bounding boxes for the re-ID network training.

The end-to-end methods jointly train the person detection and re-ID networks. Xiao~\textit{et al.}~\cite{cuhk} firstly proposed an end-to-end person search network and provided a benchmark CUHK-SYSU dataset. Chen~\textit{et al.}~\cite{chen2020hoim} employed the background features as negative samples to train the re-ID network. Chen~\textit{et al.}~\cite{norm_aware} separated the feature embedding into the norm and angle which are used as a detection confidence score and an identity feature, respectively.
Zhang~\textit{et al.}~\cite{zhang2021diverse} pretrained an external re-ID network which is then used as a strong teacher model to supervise the re-ID network based on the knowledge distillation framework. Li and Miao~\cite{seqnet} employed an additional Faster R-CNN header sequentially to extract the superior identity features from the high-quality person proposals. Han~\textit{et al.}~\cite{han2021end} adaptively controlled the gradient backpropagation to train the sub-networks of the re-ID and part classification according to the quality of detection results. Lee~\textit{et al.}~\cite{lee2022oimnet} suggested a feature standardization scheme and a localization aware memory updating scheme to alleviate the effect of class imbalance and inaccurately detected proposals, respectively.
The transformer architectures were also employed to improved the performance of person search~\cite{cao2022pstr},~\cite{yu2022cascade},~\cite{fiaz2023sat},~\cite{yang2024unleashing}. 
Recently, Oh~\textit{et al.}~\cite{oh2024domain} assumed the training data of real target domains are not available, and proposed a domain generalizable person search method that uses only an unreal dataset for training.

On the other hand, the weakly-supervised person search has been introduced that uses the labeled bounding boxes only without using the identity labels for training~\cite{han2021weakly},~\cite{han2021context},~\cite{yan2022exploring},~\cite{wang2023self}. Moreover, domain-adaptation methods have been proposed to address the unsupervised person search problem where both of the bounding box and identity labels are not available~\cite{daps, almansoori2024ddam}. Note that the existing methods of person search have been usually developed considering a single target dataset only, and hence suffer from the catastrophic forgetting problem where new target datasets are continuously given in the lifelong learning scenario.

\begin{figure*}[t]
	\centering
	\includegraphics[width=0.99\textwidth]{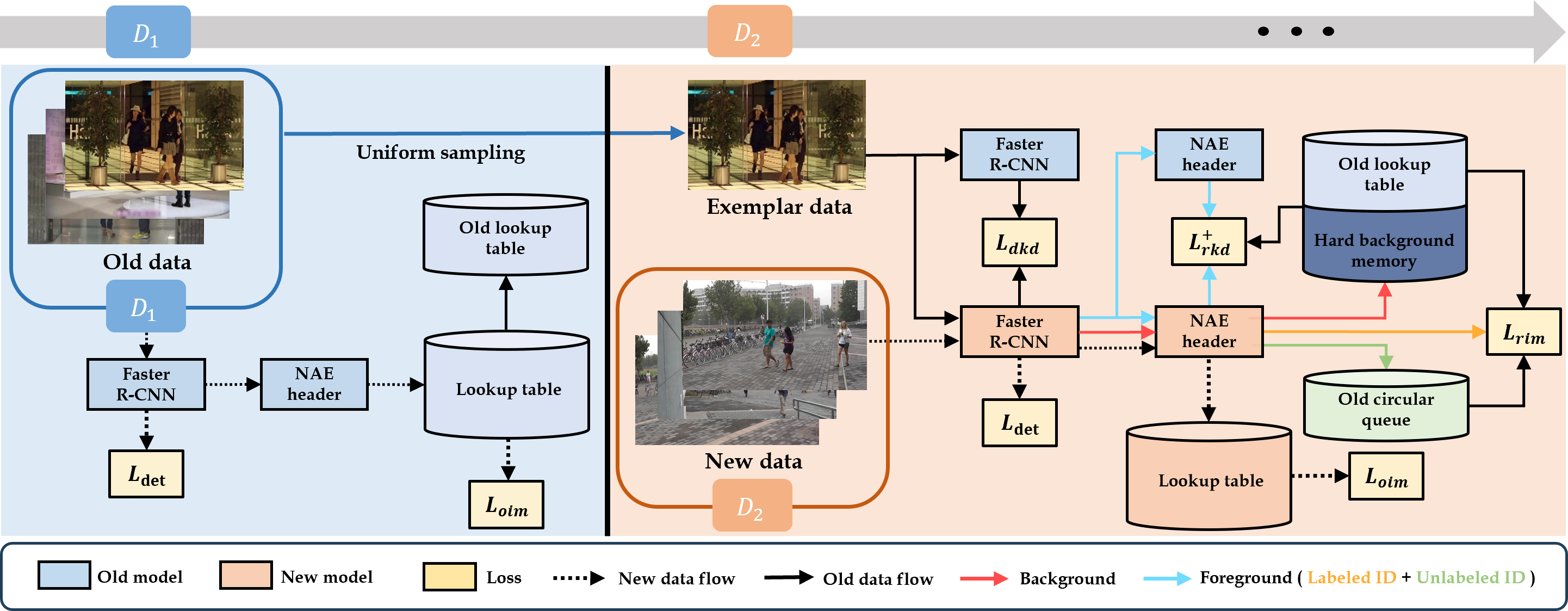}
	\caption{Overall framework of the proposed method.}
	\label{fig:overall_framework}
  \end{figure*}

\subsection{Lifelong Object Detection}
Lifelong object detection methods are classified into the class-incremental approach and the domain-incremental approach. 
The class-incremental object detection considers a certain target dataset where the new object classes are incrementally added. Shmelkov~\textit{et al.}~\cite{shmelkov2017incremental} first introduced the problem of catastrophic forgetting in the object detection. Adaptive distillation has been performed between the intermediate features and the output of the region proposal network based on the end-to-end framework~\cite{hao2019end,peng2020faster}. Shieh~\textit{et al.}~\cite{shieh2020continual} stored a subset of old data into the exemplar memory to alleviate the catastrophic forgetting in old classes. Liu~\textit{et al.}~\cite{liu2023continual} focused on the most informative old knowledge by sampling the most reliable foreground prediction from the old model, which are then used as pseudo labels in a transformer based detection network, DETR~\cite{carion2020end}. Dong~\textit{et al.}~\cite{dong2023incremental} performed self-supervised learning with the DETR network where only a few labeled new object classes appear in the new data.
On the other hand, object detection datasets are associated with different domains according to the variations of background, lighting, and camera viewpoint, even though they contain the same object classes. The domain-incremental object detection assumes incrementally added new domains with the same object class. Li~\textit{et al.}~\cite{li2022technical} used a transformer-based feature extractor to adaptively apply the classification head network to each newly added domain. Mirza~\textit{et al.}~\cite{mirza2022efficient} stored the statistical changes across the domains used to perform the task on the corresponding specific domain. 

\subsection{Lifelong Person Re-ID}
Wu and Gong~\cite{wu2021generalising} first introduced the lifelong person re-ID problem and performed coherence learning for classification, distribution, and representation, respectively. Pu~\textit{et al.}~\cite{pu2021lifelong} adaptively accumulated the knowledge of old domains via the instance-based similarity to improve the generalization ability. Ge~\textit{et al.}~\cite{ge2022lifelong} developed a domain adaptation framework that reduces the gap between the old and new domains by using the augmented new data following the distributions of the old domains. Sun and Mu~\cite{sun2022patch} selected diverse and important patches from images by using a differentiable patch sampler to preserve both the local and global relational knowledge. Huang~\textit{et al.}~\cite{huang2022lifelong} developed a relation consistency learning to encourage the new model to return the consistent results of similarity ranking to that of the old model. Yu~\textit{et al.}~\cite{yu2023lifelong} proposed a knowledge transfer scheme via the bi-directional learning that dynamically updates the old model while training the new model. Xu~\textit{et al.}~\cite{xu2024lstkc} used a pair-wise relation matrix to filter out the erroneous knowledge of the old model and transfer the refined knowledge to the new model.

\section{Proposed Method}

Fig.~\ref{fig:overall_framework} shows the overall architecture of the proposed end-to-end LPS framework. We use the SeqNet~\cite{seqnet} as a baseline network which consists of the Faster R-CNN~\cite{ren2015faster} and the NAE (norm-aware embedding)~\cite{norm_aware} header. 
Let us assume that a sequence of person search datasets in different domains are given in order, as $D_1 \rightarrow D_2 \rightarrow \ldots \rightarrow D_N$. The model is trained by using the first dataset $D_1$.
When the new dataset $D_2$ is given, we regard the model trained on $D_1$ as the old model, and construct a new model by replicating the old model.
Then the new model is trained by using $D_2$ and a small subset of $D_1$, called exemplar data, to avoid the knowledge forgetting of $D_1$. We also use the representative features of person identities, called prototypes~\cite{rebuffi2017icarl}, stored in the old look-up table (LUT) $\mcalT$.
Whenever a new dataset $D_N$ is available, the old model is replaced with the new model, and the new model is re-trained by using the new data $D_N$ as well as both the old exemplar data and the old prototypes selected from $\left\{D_1, \cdots, D_{N-1}\right\}$ to mitigate the catastrophic forgetting.
We use a small subset of the old data following the typical rehearsal (replay) based methodology of lifelong learning~\cite{rebuffi2017icarl,shieh2020continual, wu2021generalising, ge2022lifelong}. 
However, it is worth to note that we do not employ multiple old models but always have a single old model which is updated whenever a new dataset is given. The old model conveys the knowledge of the previous domains and thus the parameters of the old model are frozen during the training of the new model.
We preserve the knowledge of the old data while training the model using the new data via knowledge distillation between the old and new models.

\begin{figure*}[t]
	\centering
	\includegraphics[width=0.9\linewidth]{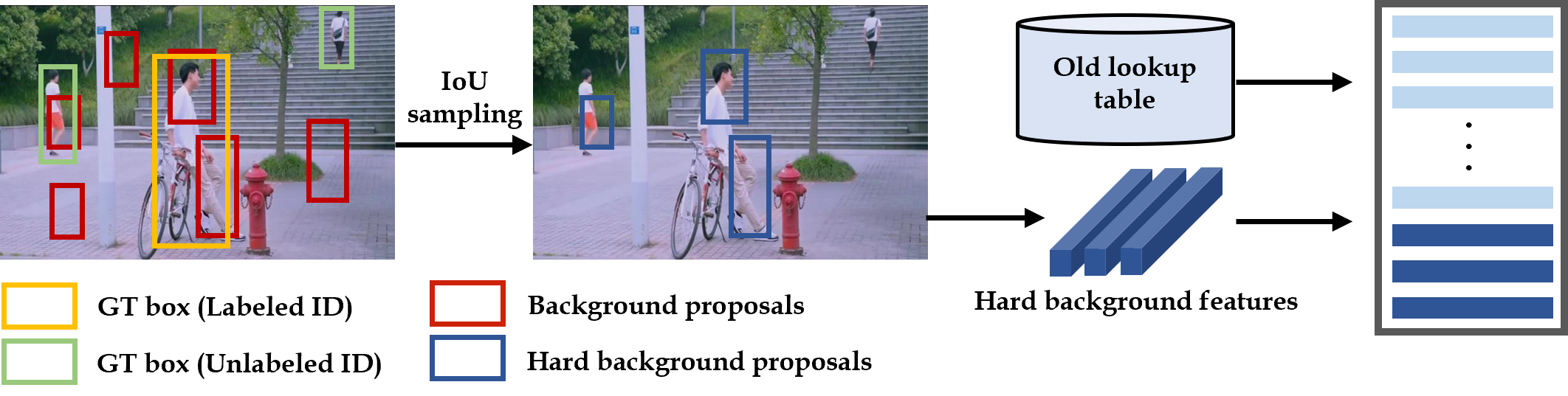}
	\caption{Sampling of the hard background proposals detected from the exemplar data.}
	\label{fig:exemplar_det_results}
\end{figure*}

\subsection{Re-ID Knowledge Distillation}
\subsubsection{Prototype-Based Distillation}
Existing methods of lifelong person re-ID~\cite{wu2021generalising, ge2022lifelong} perform the knowledge distillation by matching the distributions of the feature similarity between the old and new models within a mini-batch. However, it may not provide faithful results when applied to LPS, since a mini-batch is composed of relatively small numbers of scene images, and small numbers of identities accordingly, due to the memory constraints. Furthermore, multiple person instances with the same identity are rarely included in a single mini-batch according to the uniqueness prior~\cite{han2021context}. 

To address this issue for LPS, we utilize the stored prototypes of person identities as informative guidance for the re-ID knowledge distillation.
The prototype is computed by aggregating the features of diverse person instances with the same identity~\cite{cuhk, norm_aware, seqnet, yu2022cascade}. 
Let $\mcalF$ denotes the set of the foreground proposals in the exemplar data of a single mini-batch, that is detected by the Faster R-CNN of the new model. Let $\bx^{\text{old}}_i$ and $\bx^{\text{new}}_i$ be the L2-normalized features of the $i$-th proposal in $\mcalF$, that are extracted through the NAE headers of the old and new models, respectively. We estimate the target distribution of the feature similarity of $\bx^{\text{old}}_i$ compared to all the prototypes in $\mcalT$, such that the probability $q_{i,k}$ associated with $\bx^{\text{old}}_i$ and $\bz_k$, the $k$-th prototype in $\mcalT$, is given by 
\begin{equation}
   q_{i,k} = \frac{\exp \left(\bz_k^{\mathrm{T}}\bx^{\text{old}}_i  / \tau_d\right)}{\sum_{\bz \in \mcalT} \exp \left(\bz^{\mathrm{T}}\bx^{\text{old}}_i / \tau_d\right)}.
  \label{eq:target_distribution}
\end{equation}
We also estimate the predicted distribution of the feature similarity of $\bx^{\text{new}}_i$ compared to all the prototypes, such that the probability $p_{i,k}$ associated with $\bx^{\text{new}}_i$ and $\bz_k$ is given by
\begin{equation}
  p_{i,k} = \frac{\exp \left(\bz_k^{\mathrm{T}}\bx^{\text{new}}_i  / \tau_d\right)}{\sum_{\bz \in \mcalT} \exp \left(\bz^{\mathrm{T}}\bx^{\text{new}}_i / \tau_d\right)}.
  \label{eq:pred_distribution}
\end{equation}

Note that the predicted similarity distribution with respect to the prototypes in the old domains changes when the model is trained on the new domain, which could cause the forgetting of the re-ID knowledge learned on the old domains. Therefore, we train the new model to generate a more consistent distribution to the target distribution by employing a prototype-based re-ID knowledge distillation loss given by
\begin{equation}
  \mathcal{L}_{\text{rkd}} = \frac{1}{|\mcalF|} \frac{1}{|\mcalT|} \sum_{i \in I(\mcalF)} \sum_{k \in I(\mcalT)}  q_{i,k} \log \frac{q_{i,k}}{p_{i,k}},
  \label{eq:prot}
\end{equation}
where $I(\cdot)$ means the index set. 
By minimizing $\mathcal{L}_{\text{rkd}}$, the new model is trained to yield a more consistent distribution to the target distribution with respect to the prototypes in the old domains, and eventually extracts unique and representative features of all person identities alleviating the catastrophic forgetting in re-ID.  \\

\subsubsection{Hard Background Proposal-Based Distillation}
Though the prototypes in the old LUT serve as a good prior for the re-ID knowledge distillation, we further improve the performance by using the background proposals in the old domains additionally. At each iteration, the new model detects the background proposals from the scene images in the exemplar data, as depicted in the red boxes in Fig.~\ref{fig:exemplar_det_results}. Inaccurate background proposals are often generated that partially overlap with the foreground person instances. We refer them as the hard background proposals. The hard background proposals convey the partial information of the person identities, exploited to improve the discrimination performance of the person identities.

Specifically, we sample the hard background proposals that have the higher intersection over union (IoU) scores than a certain threshold $\lambda_{\rm b}$, with respect to the ground truth bounding boxes of the foreground persons, as depicted in the blue boxes in Fig.~\ref{fig:exemplar_det_results}. Then we store the re-ID features of the hard background proposals into the feature memory $\mcalM$.
We re-compute the distributions of the feature similarity compared to all the prototypes in $\mcalT$ as well as all the features of the hard background proposals in $\mcalM$, such that the probabilities $q_{i,k}^+$ and $p_{i,k}^+$ associated with $\bz_k$, the $k$-th element in $\mcalT \cup \mcalM$, are given by 
\begin{equation}
	q_{i,k}^+ = \frac{\exp \left(\bz_k^{\mathrm{T}} \bx^{\text{old}}_i / \tau_d\right)}
	{\sum_{\bz \in \left\{\mcalT \cup \mcalM\right\}} \exp \left(\bz^{\mathrm{T}}\bx^{\text{old}}_i / \tau_d\right)},
	\label{eq:hard_bg_target_distribution}
\end{equation}
\begin{equation}
	p_{i,k}^+ = \frac{\exp \left(\bz_k^{\mathrm{T}}\bx^{\text{new}}_i / \tau_d\right)}
	{\sum_{\bz \in \left\{\mcalT \cup \mcalM\right\}} \exp \left(\bz^{\mathrm{T}}\bx^{\text{new}}_i / \tau_d\right)}.
	\label{eq:hard_bg_pred_distribution}
\end{equation}

Accordingly, we have the refined re-ID knowledge distillation loss as
\begin{equation}
  \mathcal{L}_{\text{rkd}}^+ = \frac{1}{|\mcalF|} \frac{1}{|\mcalT \cup \mcalM|} \sum_{i \in I(\mcalF)} \sum_{k \in I(\mcalT \cup \mcalM)}  q^+_{i,k} \log \frac{q^+_{i,k}}{p^+_{i,k}}.
	\label{eq:prot}
\end{equation}
Consequently, we improve the discrimination performance of the person identities by exploiting more rich information carried by the hard background proposals. Furthermore, the features learned by additionally using the hard background proposals are more robust against the inaccurately detected person proposals, which alleviates the task conflict problem between the detection and re-ID even when the detection knowledge in the old domains is forgotten.

\subsection{Rehearsal-Based Instance Matching}
The foreground and background proposals extracted from the exemplar data are used for consistent learning between the old and new models via the re-ID knowledge distillation. Note that, as depicted in the green boxes in Fig.~\ref{fig:exemplar_det_results}, some foreground person instances have no identity labels. Such unlabeled instances can also serve as the negative samples for all the labeled identities to learn the discriminative feature representations. At the same time, the new model should be guided to minimize the feature discrepancy across the person instances in the exemplar data that have the same identity. Therefore, we also utilize the unlabeled instances in the exemplar data to further capture the re-ID knowledge in the old domains while the model is trained on the new data. 

\begin{table*}[t]
	\centering
	\caption{Statistics of person search datasets.}
	\renewcommand{\arraystretch}{1.2}
	\begin{tabular}{c|ccc|ccc}
		\toprule%
		\multirow{2}{*}{Dataset}    
		&\multicolumn{3}{c|}{Training} &\multicolumn{3}{c}{Test} \\
		\cline{2-7}
		& MovieNet-PS  &  CUHK-SYSU   & PRW & MovieNet-PS & CUHK-SYSU   & PRW   \\ 
		\midrule
		Frame     &  20,158  & 11,206 & 5,704 & 43,640  & 6,978 & 6,112 \\ 
		Identity &  2,078    & 5,532  & 483  & 1,000  & 2,900 & 544 \\ 
		Bounding box & 32,927  & 55,272 & 18,048 & 79,607  & 40,871 & 25,062 \\ 
		Query      & - & -  & - & 1,000 &  2,900  & 2,057  \\ 
		\bottomrule
	\end{tabular}
	\label{tab:ps_datasets}
\end{table*}

The features of the unlabeled proposals are stored in the old circular queue $\mcalQ$. Let $\mcalF_{\rm L}$ denote the set of the labeled proposals in $\mcalF$, and let $\bx^{\text{new}}_i$ be the feature of the $i$-th proposal in $\mcalF_{\rm L}$ extracted by the new model. We compute the probability that $\bx^{\text{new}}_i$ is classified into its ground truth label as
\begin{equation}
    \rho_{i} = \frac{\exp \left(\bz(i)^{\mathrm{T}} \bx^{\text{new}}_{i}  /\tau_r \right)}{ \sum_{\bz \in \mcalT}  \exp  \left(\bz^{\mathrm{T}} \bx^{\text{new}}_{i} / \tau_r\right)  +  \sum_{\by \in \mcalQ}  \exp  \left(\by^{\mathrm{T}} \bx^{\text{new}}_{i}  / \tau_r\right)}, 
    \label{eq:roim_score}
\end{equation}
where $\bz(i)$ means the prototype of the ground truth identity of the $i$-th proposal in $\mcalF_{\rm L}$, and $\by$ denotes the feature of the unlabeled proposals stored in $\mcalQ$. We train the new model to increase the classification score of the extracted features by employing the rehearsal-based instance matching loss given by
\begin{equation}
    \mathcal{L}_{\text{rim}} = -\frac{1}{|\mcalF_L|} \sum_{i \in I(\mcalF_L)} \log \rho_{i}.
\end{equation}
By minimizing $\mathcal{L}_{\text{rim}}$, we reduce the feature discrepancy between the labeled proposal and its ground truth identity while preserving the discrimination performance in the old domains with the help of the unlabeled proposals. 
It is worth to note that the conventional OIM~\cite{cuhk} loss considers the labeled identities and the unlabeled instances in a single target domain only. On the contrary, the proposed rehearsal-based loss $\mathcal{L}_{\text{rim}}$ employs the labeled identities and the unlabeled instances across the old data, aiming to preserve the discrimination performance in the old domains for lifelong learning purpose.

\subsection{Training and Inference}
At the training phase, both of the person detection and re-ID networks are trained in the end-to-end manner. Note that the baseline network of SeqNet~\cite{seqnet} also uses the losses of $\mathcal{L}_{\text{det}}$ and $\mathcal{L}_{\text{oim}}$ when training the new model by using the new dataset. To preserve the knowledge of the old domains in terms of the person detection, we additionally use the detection knowledge distillation loss ${\mathcal{L}_{\text{dkd}}}$ of the existing lifelong object detection method~\cite{peng2020faster}.
Finally, the total loss function is given by 
\begin{equation}
    \mathcal{L}_{\text{total}} = {\mathcal{L}_{\text{dkd}}} + \mathcal{L}_{\text{rkd}}^+  + \mathcal{L}_{\text{rim}} + \mathcal{L}_{\text{det}} + \mathcal{L}_{\text{oim}}.
\end{equation}

After the new model is trained with the last dataset $D_N$,  we discard the old model and only utilize the new model to detect the bounding boxes of the person instances and extract the re-ID features for all the datasets $\left\{D_1, D_2 \ldots D_N\right\}$ at the inference phase.

\section{Experimental Results}

\subsection{Experimental Setup}
\subsubsection{Datasets}
We used the three datasets to evaluate the performance of the proposed lifelong person search method. CUHK-SYSU~\cite{cuhk} and PRW~\cite{prw} are widely used for the person search task. The CUHK-SYSU dataset includes the images obtained from the street snapshots and movies, with the annotations of the bounding boxes and person identities. We set the gallery size to 100. The PRW dataset is composed of the video frames capturing a university campus by six different cameras. We also use a recently released large-scale person search dataset of MovieNet-PS~\cite{qin2023movienet} gathered from the 385 movie sequences. The MovieNet-PS is a challenging dataset since it includes the scene images with diverse backgrounds, illuminations, and poses of persons to reflect more realistic and challenging scenarios of person search. Moreover, there are many persons partially appearing due to the occlusion, and the person instances with the same identity often wear different clothes. The statistics of the three datasets are shown in Table~\ref{tab:ps_datasets}.

\renewcommand{\arraystretch}{1.0}
\begin{table*}[t]
	\small
	\centering
	\caption{Performance of the lifelong learning evaluated on the three person search datasets with the training order of CUHK-SUSU $\rightarrow$ PRW $\rightarrow$  MovieNet-PS. The performance on each dataset is measured by using the model after the training with the last dataset is over. The best scores are boldfaced.}
	\resizebox{\textwidth}{!}{%
		\renewcommand{\arraystretch}{1.3}
		\begin{tabular}{c|cc|cc|cc|cc|cc|cc|cc|cc}
			\toprule
			\multirow{3}{*}{Methods}                                                                                                
			& \multicolumn{4}{c|}{CUHK-SYSU}  & \multicolumn{4}{c|}{PRW}& \multicolumn{4}{c|}{MovieNet-PS}& \multicolumn{4}{c}{Average}               \\
			\cline{2-17}
			& \multicolumn{2}{c|}{Detection}  & \multicolumn{2}{c|}{Re-ID} & \multicolumn{2}{c|}{Detection}  & \multicolumn{2}{c|}{Re-ID} & \multicolumn{2}{c|}{Detection}  & \multicolumn{2}{c|}{Re-ID} & \multicolumn{2}{c|}{Detection}  & \multicolumn{2}{c}{Re-ID} \\
			\cline{2-17}
			& \multicolumn{1}{c}{Recall} & AP & mAP&Top-1& Recall & AP & mAP&Top-1& Recall & AP & mAP&Top-1& Recall & AP & mAP&Top-1\\
			\cmidrule(r){1-17}
			Joint-Train  & 91.3 & 88.1 & 92.9 & 93.7 & 97.8 & 94.8 & 50.9 & 85.4  & 87.1 & 82.9 & 39.9  & 81.2 & \multicolumn{1}{c}{92.1} & 88.6 & 61.2 &  86.8                    \\ \midrule    \midrule
			FineTune        & 55.0  & 52.3  & 73.1  & 75.1 & 61.1 & {60.0} & 11.7 & 57.2 & \textbf{91.4} & \textbf{87.1} & \textbf{40.1} & \textbf{82.3} & 69.2 & 66.5 & 41.6 & 71.5                     \\ \midrule
			Det + AKA & 79.5 & 77.9 & 66.0 & 65.1 & 91.9 & 89.6 & 7.4  & 31.2 & 82.3 & 76.1  & 13.0 & 49.6 & 84.6 & 81.2 & 28.8 & 48.6                     \\
			Det + PTKP   & 79.5 & 77.9 & 82.2 & {83.9} & 91.9 & {89.6} & 33.2 & 72.9 & 82.3 & 76.1 & 20.0 & 66.8 & 84.6 & 81.2 & 45.1 &  74.5                   \\
			Det + KRKC   & 79.5 & 77.9 & 83.3 & 85.9 & 91.9 & {89.6} & 29.3 & 80.7 & 82.3 & 76.1 & 25.2 & 75.9 & 84.6 & 81.2 & 45.9 & 80.8                    \\
			\midrule
			\textbf{Proposed}       & \textbf{81.4} & \textbf{79.7}  & \textbf{91.2} & \textbf{92.6} & \textbf{93.9} & {\textbf{91.3}} & \textbf{42.2}  & \textbf{83.2} & 85.1 & 78.4  & 34.1 & 78.6 & \textbf{86.8} & \textbf{83.1} & \textbf{55.8}  & \textbf{84.8}                     \\
			\midrule \midrule
			Det* + AKA   & 100 & 100 & 67.2 & 65.8 & 100 & 100 & 7.7 & 31.3 & 100 & 100 & 15.5 & 53.6 & 100 & 100  & 30.1 & 50.2                       \\
			Det* + PTKP   & 100 & 100 & 86.1 & 86.4 & \multicolumn{1}{c}{100} & 100 & 35.8 & 74.1 & \multicolumn{1}{c}{100} & 100 & 23.4 & 70.3 & \multicolumn{1}{c}{100} & 100  & 48.4 & 76.9                    \\
			Det* + KRKC   & 100 & 100 & 88.1 & 89.9 & \multicolumn{1}{c}{100} & 100 & 31.3 & 82.6 & \multicolumn{1}{c}{100} & 100 & 29.1 & 79.9 & \multicolumn{1}{c}{100} & 100  & 49.5 &   84.1                \\
			\midrule
			\textbf{Proposed*}       & \textbf{100} &  \textbf{100} & \textbf{92.8} & \textbf{93.6} & \textbf{100} & \textbf{100} & \textbf{43.4}  & \textbf{84.4} & \textbf{100} & \textbf{100}  & \textbf{41.3} & \textbf{84.4}  & \textbf{100} & \textbf{100} &  \textbf{59.2} & \textbf{87.4}                     \\
			\bottomrule 
		\end{tabular} 
	}
	\label{tab:incremental_kompare_order1}
\end{table*}

\subsubsection{Evaluation Metrics}
We evaluated the performance of the person detection and re-ID, respectively, based on the LPS framework. We used the recall and the average precision (AP) to measure the detection performance. The recall calculates the percentage of the true positive bounding boxes where the IoU scores with respect to any ground truth bounding box are higher than 0.5.
The AP computes the average precision of the bounding boxes by measuring the area under the Precision-Recall curve using the IoU scores with respect to the ground truth.
We also used the mean Average Precision (mAP) and the Top-$k$ scores for the re-ID. The mAP calculates the averaged precision of searching a query from the gallery images. It measures the area under the Precision-Recall curve using the feature similarities between the query and all the detected gallery persons. The detected bounding boxes that overlap with the ground-truth with the IoU scores higher than 0.5 are set as the true positives. The Top-$k$ score checks whether at least one of the $k$-most similar candidates is a true positive or not. We adopted the Top-1 score in this work.

\subsubsection{Implementation Details}
We uniformly sampled 2\% of the training data from the old datasets to construct the exemplar data, as did in many literatures of lifelong person re-ID~\cite{wu2021generalising,ge2022lifelong, yu2023lifelong}. 
We store the prototypes of the persons in the exemplar data only into the old LUT, which are not updated during the training on the new domain.
For the model training, we set the batch size for the exemplar data to 2 for each old domain and 5 for the new domain, respectively. We resized the input image to 1500 $\times$ 900 and applied the random horizontal flipping. We set the size of the old circular queue $\mcalQ$ as 1000.
We trained the model until it reaches the highest performance on the first domain, and trained the model for 5 epochs each on the other domains. Since our baseline setting achieves the best performance when the model is trained during 5 epochs for each new dataset, we also trained the model for 5 epochs on the new dataset for fair comparison to the baseline setting.
The initial learning rate is set to 0.003, which is warmed up with a learning rate scheduler and further decayed by the value of 0.1 in the 3rd epoch of each new domain. For the stochastic gradient descent, we set the momentum value to 0.9 and the weight decay to 0.0005. $\lambda_{\rm b}$, $\tau_d$, and $\tau_r$ are empirically set to 0.1, 0.3 and 0.1, respectively. All our experiments were implemented using PyTorch and a single NVIDIA TITAN X GPU.

\begin{figure}[t]
	\centering
	\begin{minipage}[b]{0.49\linewidth}
		\centering
		\includegraphics[width=\linewidth]{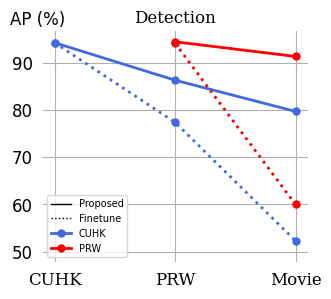}
	\end{minipage}
	\begin{minipage}[b]{0.49\linewidth}
		\centering
		\includegraphics[width=\linewidth]{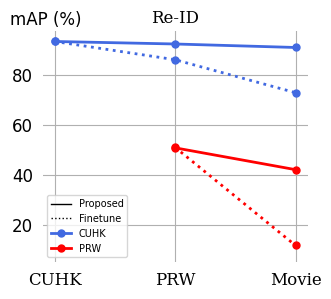}
	\end{minipage}
	\caption{Performance in terms of the old knowledge forgetting in LPS evaluated on the old datasets of CUHK-SYSU (blue) and PRW (red), when the model is sequentially trained on each new dataset in the $x$-axis. The performance of the FineTune method and the proposed method are shown with the solid and dashed lines, respectively.}
	\label{fig:forggeting_plot}
\end{figure}

\subsection{Lifelong Learning Performance}
We evaluate the lifelong learning performance of the proposed method with the training order of CUHK-SYSU $\rightarrow$ PRW $\rightarrow$ MovieNet-PS in Table~\ref{tab:incremental_kompare_order1}. We first compare two different methods: Joint-Train and FineTune, implemented on our baseline network. The Joint-Train method trains the model by using all the available datasets simultaneously. The FineTune method trains the model on each dataset in order, where the model is initialized with the previously trained weights and then re-trained by using the new dataset. At the inference phase, the performance is evaluated on every dataset by using the model trained on the last dataset.

The Joint-Train method achieves the best performance since it uses all the training datasets at once, however, it requires a huge burden of the computation as well as the storage space. We observe that the proposed method provides comparable results to the Joint-Train method and outperforms the FineTune method in terms of the averaged mAP and Top-1 score, respectively.
Note that the FineTune method sequentially re-trains the model on each dataset without using the previous old datasets, and thus its performance on the last dataset, MovieNet-PS, is relatively high. On the contrary, the proposed method uses a small subset of the old data to alleviate the knowledge forgetting and slightly degrades the performance on the last dataset compared to the FineTune method. However, the proposed method significantly outperforms the FineTune method in the old datasets of CUHK-SYSU and PRW.

Fig.~\ref{fig:forggeting_plot} shows the performance evaluated on the old datasets of CUHK-SYSU (blue) and PRW (red), when the model is sequentially trained on the new datasets in the $x$-axis, in order.
As shown by the dashed lines, both the AP and mAP scores are decreased in the FineTune method showing the knowledge forgetting effect of both the detection and re-ID tasks in the LPS scenario. However, the proposed method significantly mitigates such performance degradation as shown by the solid lines, and successfully preserves the old knowledge for LPS.

\begin{figure}[t]
	\centering
	\begin{minipage}[b]{0.49\linewidth}
		\centering
		\includegraphics[width=\linewidth]{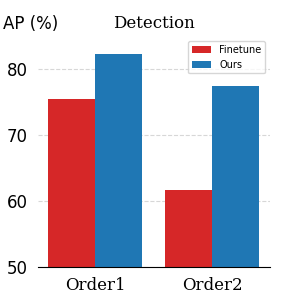}
	\end{minipage}
	\begin{minipage}[b]{0.49\linewidth}
		\centering
		\includegraphics[width=\linewidth]{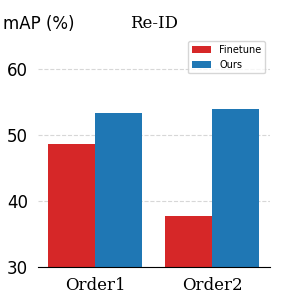}
	\end{minipage}
	\caption{Performance of the proposed method with two different training orders. Order1: MovieNet-PS $\rightarrow$ CUHK-SYSU  $\rightarrow$ PRW. Order2: PRW $\rightarrow$ CUHK-SYSU $\rightarrow$MovieNet-PS.}
	\label{fig:additional_order}
\end{figure}

In addition, we show the performance of the proposed LPS method when the model is trained with different orders of the datasets. In Fig.~\ref{fig:additional_order}, Order1 and Order2 represent the training orders of `MovieNet-PS $\rightarrow$ CUHK-SYSU  $\rightarrow$ PRW' and  `PRW $\rightarrow$ CUHK-SYSU $\rightarrow$ MovieNet-PS,' respectively. 
In both orders, we see that the proposed method still outperforms the FineTune methods in terms of the detection and re-ID performances, which indicates that the proposed method provides reliable performance regardless of the training orders.

\begin{figure*}[t]
	\large
	\centering
 
        \begin{minipage}[b]{0.24\textwidth}
		\includegraphics[height=0.104\textheight]{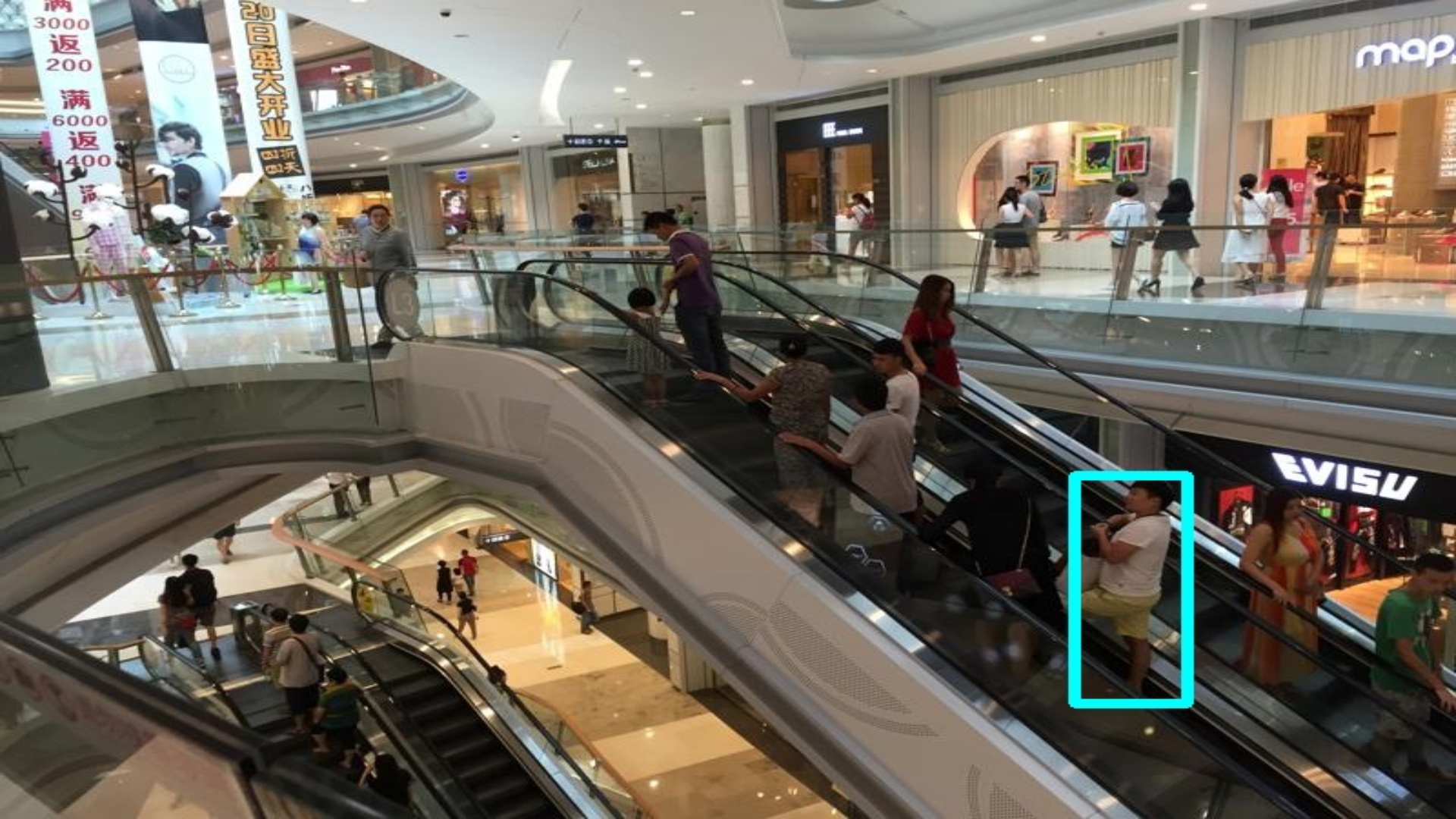}
	\end{minipage}
	\begin{minipage}[b]{0.24\textwidth}
		\includegraphics[height=0.104\textheight]{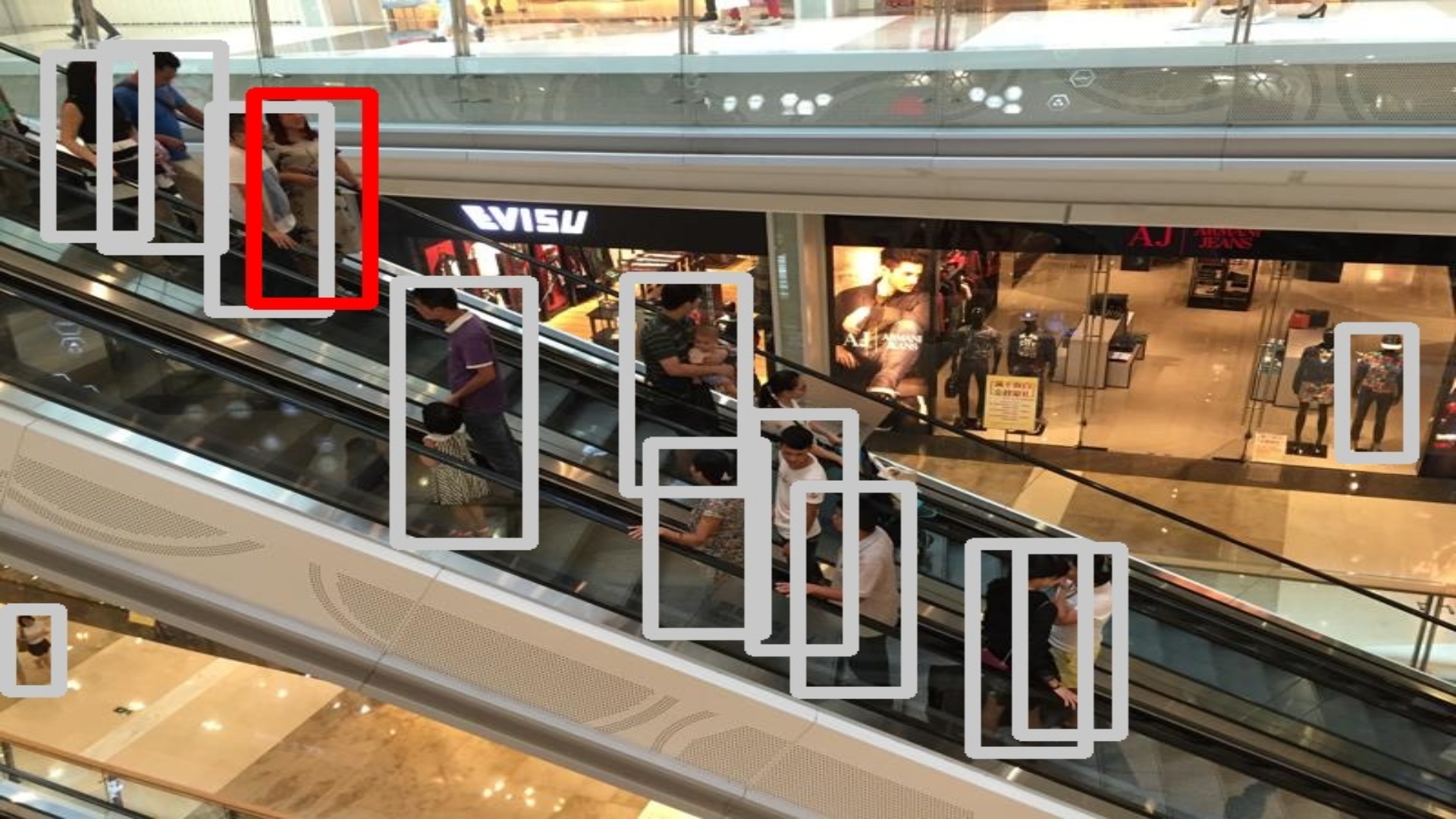}
	\end{minipage} 
	\begin{minipage}[b]{0.24\textwidth}
		\includegraphics[height=0.104\textheight]{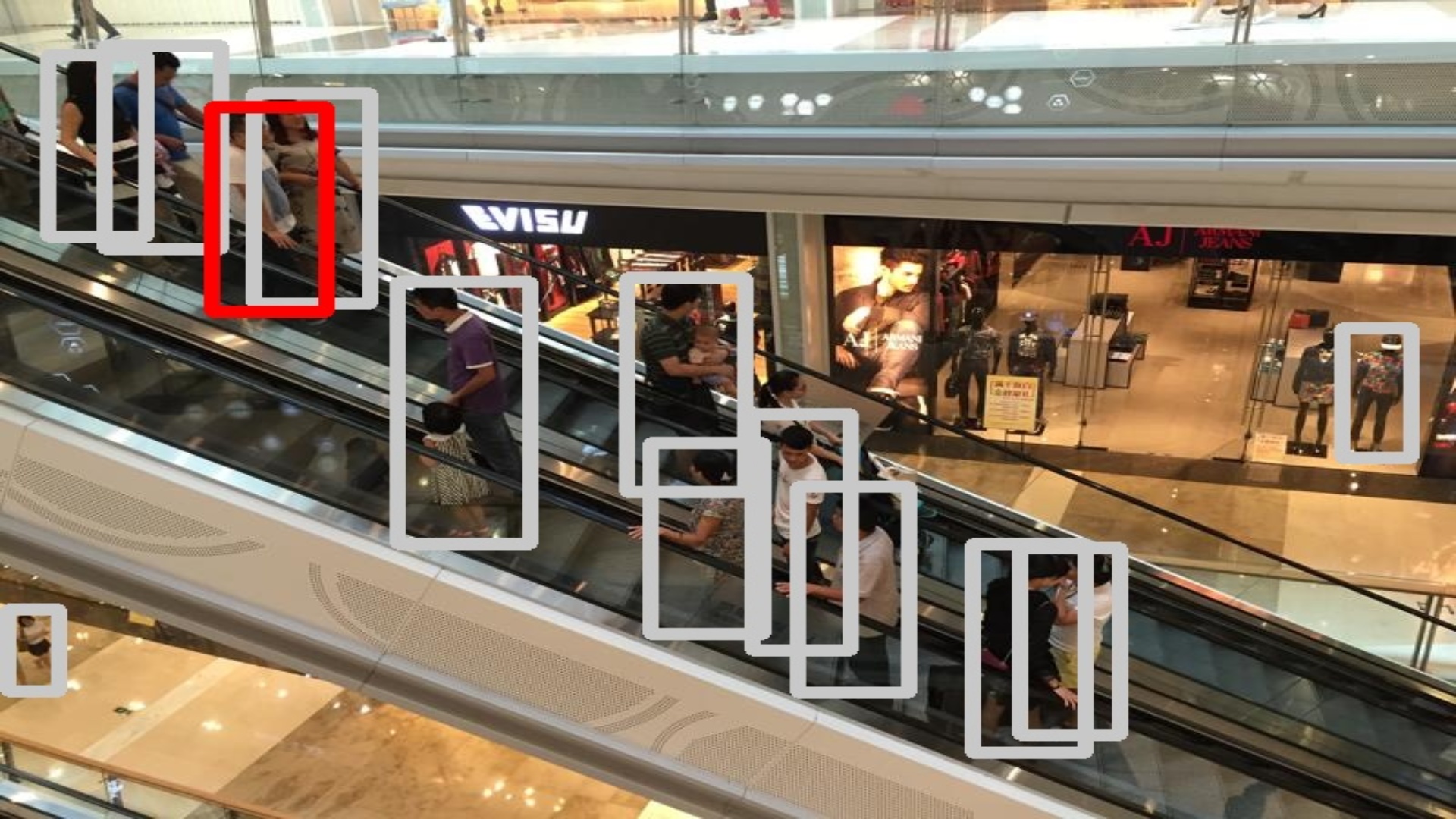}
	\end{minipage}
	\begin{minipage}[b]{0.24\textwidth}
		\includegraphics[height=0.104\textheight]{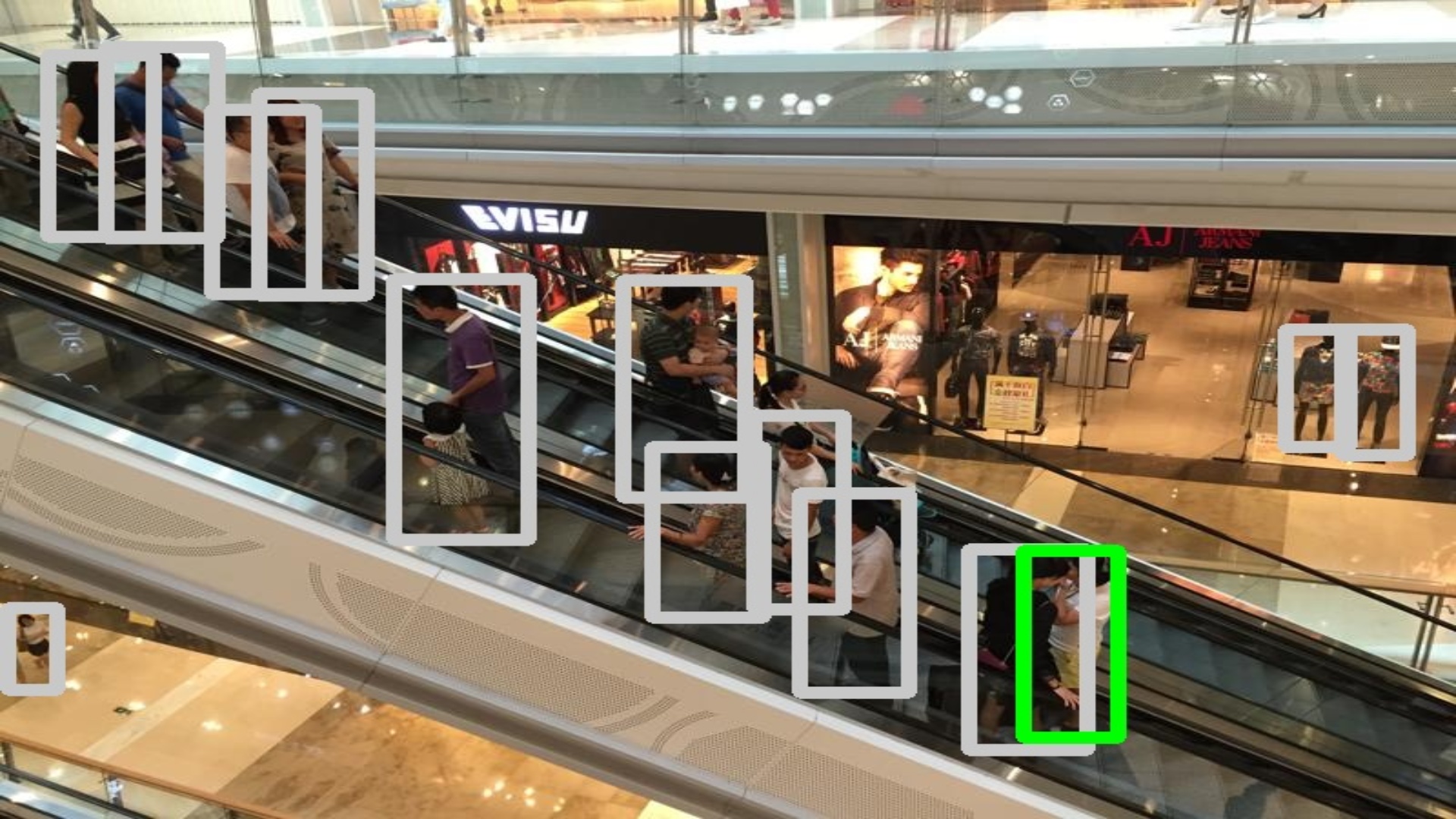}
	\end{minipage}\\  
 
	\begin{minipage}[b]{0.24\textwidth}
		\includegraphics[height=0.104\textheight]{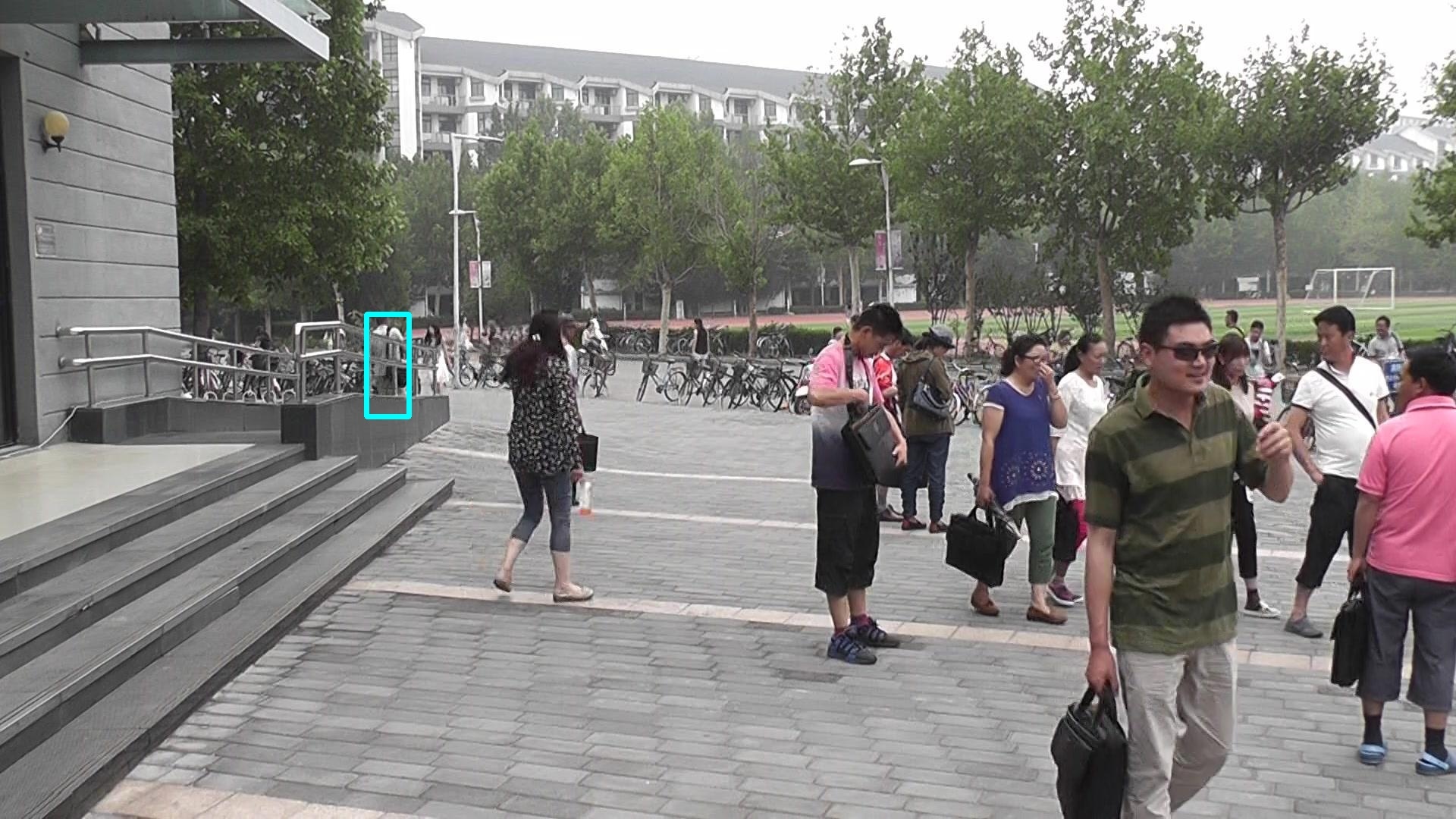}
	\end{minipage}
	\begin{minipage}[b]{0.24\textwidth}
		\includegraphics[height=0.104\textheight]{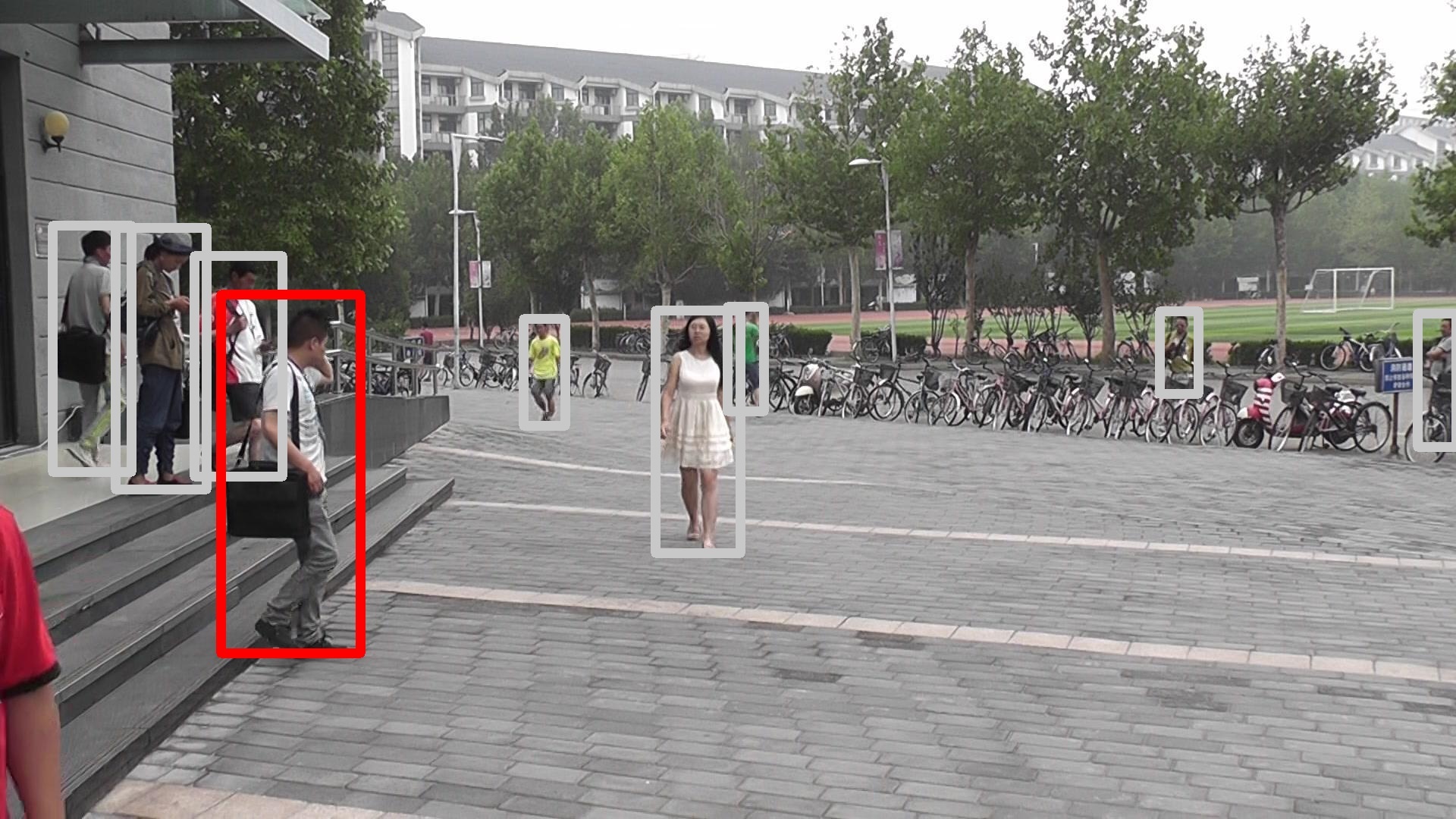}
	\end{minipage} 
	\begin{minipage}[b]{0.24\textwidth}
		\includegraphics[height=0.104\textheight]{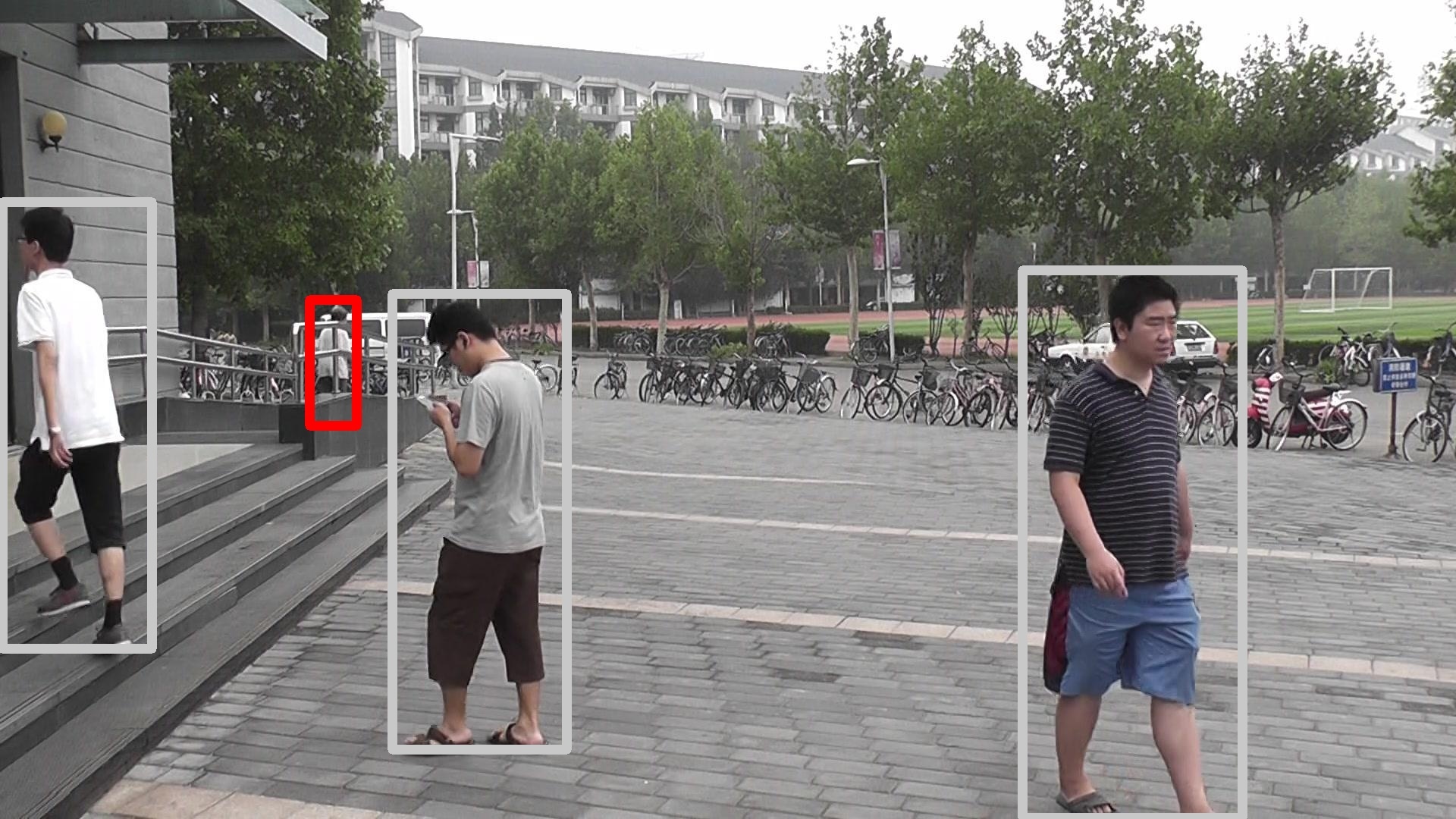}
	\end{minipage}
	\begin{minipage}[b]{0.24\textwidth}
		\includegraphics[height=0.104\textheight]{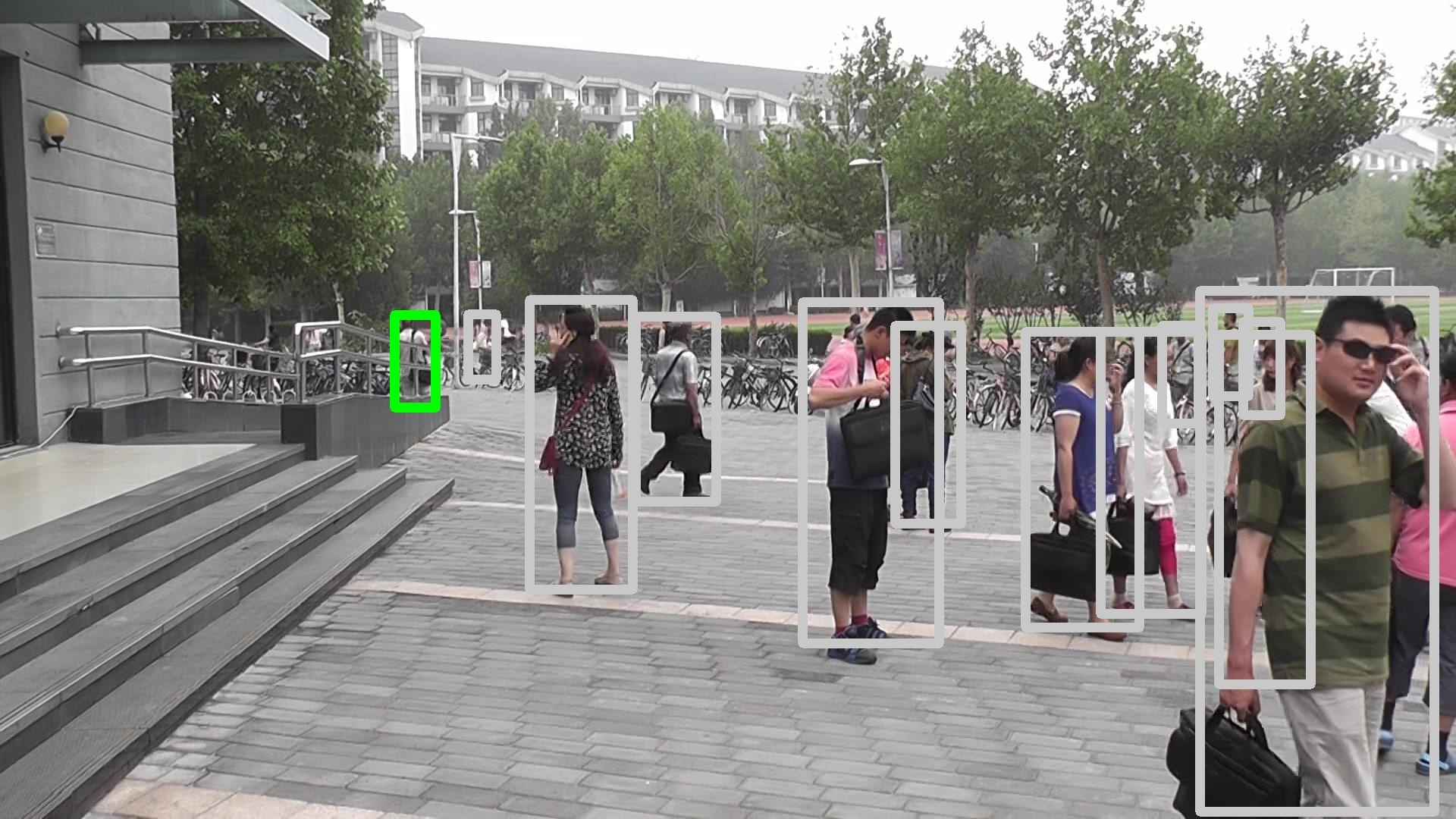}
	\end{minipage}\\

	\begin{minipage}[b]{0.24\textwidth}
		\includegraphics[height=0.104\textheight]{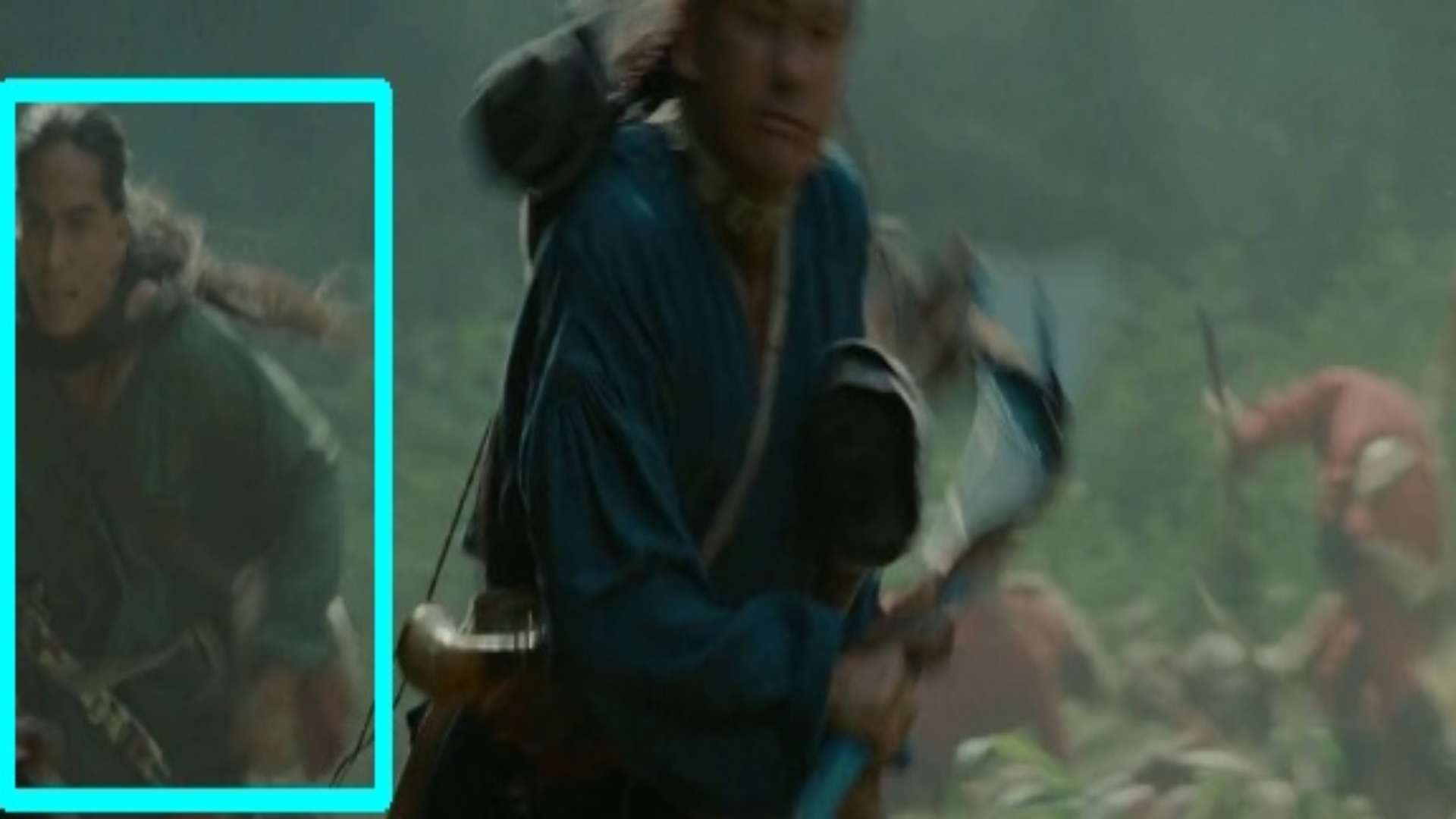}
        \caption*{(a) Query}
	\end{minipage}
	\begin{minipage}[b]{0.24\textwidth}
		\includegraphics[height=0.104\textheight]{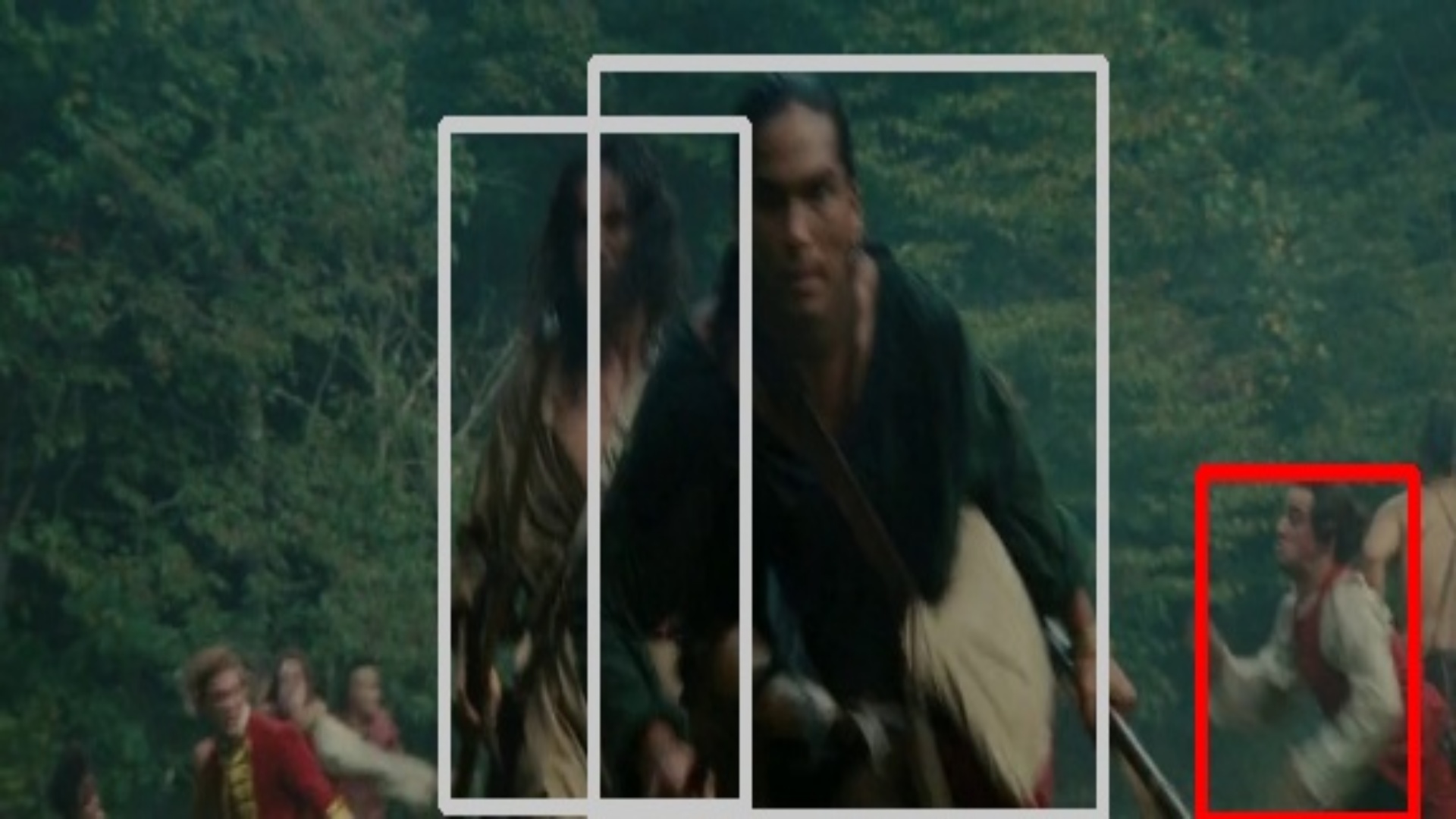}
        \caption*{(b) PTKP}
	\end{minipage} 
	\begin{minipage}[b]{0.24\textwidth}
		\includegraphics[height=0.104\textheight]{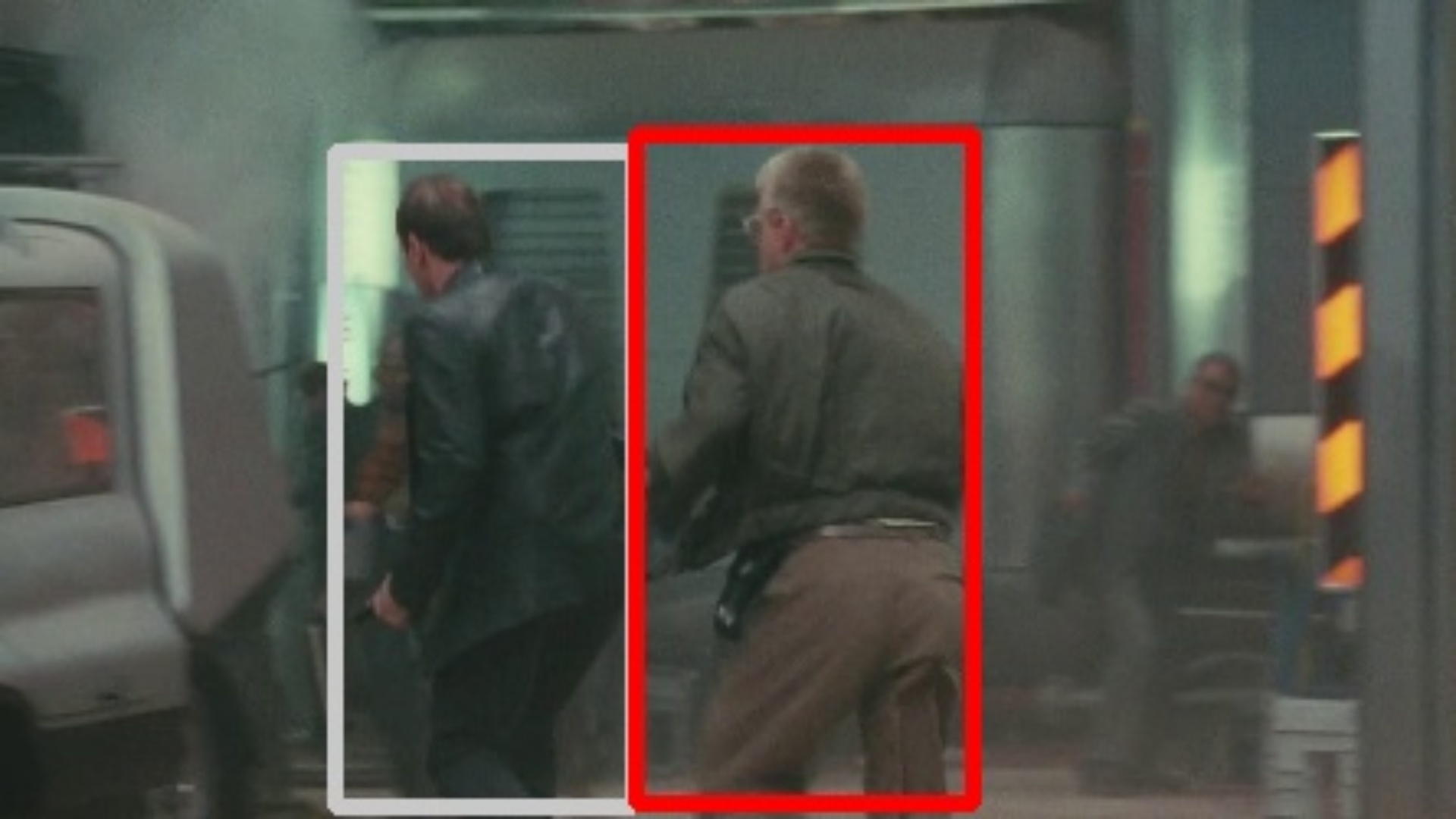}
        \caption*{(c) KRKC}
	\end{minipage} 
	\begin{minipage}[b]{0.24\textwidth}
		\includegraphics[height=0.104\textheight]{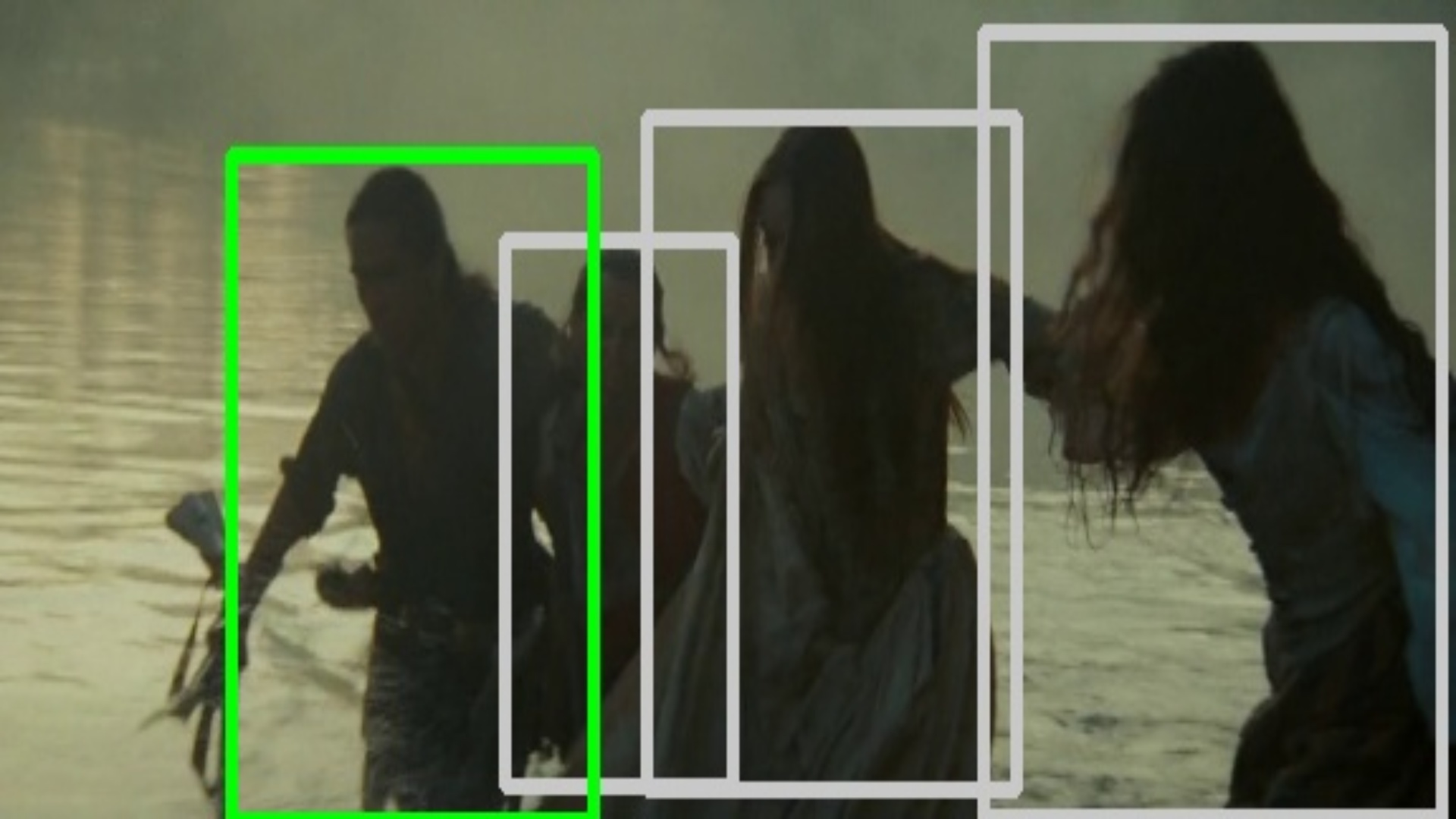}
        \caption*{(d) Proposed}
	\end{minipage} \\ 
	\caption{Qualitative comparison of the Top-1 results. From top to bottom, we show the results evaluated on the test images in CUHK-SYSU, PRW, and MovieNet-PS, respectively. The query, true positive, and false positive are depicted by the blue, green, and red boxes, respectively.}
	\label{fig:qualitative_compare}
\end{figure*}

\subsection{Comparison with Two-Step Methods}
Note that we first introduce the new problem of LPS in this paper, and there is no existing method fairly comparable to the proposed method. We attempted to conduct the additional comparative experiments by using the recent lifelong re-ID methods of AKA~\cite{pu2021lifelong}, PTKP~\cite{ge2022lifelong} and KRKC~\cite{yu2023lifelong}. The lifelong re-ID methods work on the cropped person images only and cannot be directly applied to our LPS framework that considers the scene images. We instead implemented the two-step person search framework by using the existing lifelong person re-ID methods, where the detection and re-ID networks are trained separately. We used the ResNet-50 as the backbone network for the compared methods. We also trained the detection network by using the detection knowledge distillation loss $\mathcal{L}_{\text{dkd}}$ for fair comparison. 

Table~\ref{tab:incremental_kompare_order1} compares the quantitative performance of lifelong learning where we see that the proposed method outperforms the two-step methods of `Det + AKA,' `Det + PTKP,' and `Det + KRKC' in terms of both the detection and re-ID performance. 
It means that whereas the two-step implementation of LPS by using the existing lifelong re-ID methods does not effectively reflect huge domain gaps among the person search datasets with severely different characteristics, the proposed end-to-end framework alleviates such domain gaps faithfully and is more generalizable to diverse person search datasets.
We also compared the upper-bound performance of the person re-ID by adopting the ground-truth (GT) bounding boxes for person detection, denoted as `Det* + AKA,' `Det* + PTKP,' `Det* + KRKC,' and Proposed*. Note that all the methods yield the perfect performance of detection in terms of the recall and AP, but the proposed method achieves better performance of re-ID compared with all the two-step methods.

Fig.~\ref{fig:qualitative_compare} visualizes the qualitative results of the proposed method and the two-step methods. We observe that both of PTKP~\cite{ge2022lifelong} and KRKC~\cite{yu2023lifelong} fail to find the query persons correctly in the challenging cases, for example, with the occluded persons and/or relatively small bounding boxes in the scene images. However, the proposed method successfully matches the query persons even in such cases, demonstrating the robustness to diverse LPS scenarios. 

It is worth to note that the two-step person search framework trains the backbone network when training the detection and re-ID networks, respectively, and hence requires high computational complexity and huge memory space.
In contrary, the proposed end-to-end method shares the backbone network between the jointly trained detection and re-ID networks, and is a more promising tool for LPS considering practical real-world applications.

\begin{table*}[t]
	\Large
	\centering
	\caption{Effect of the losses used in the proposed method. The performance on each dataset is measured by using the model after the training with the last dataset is over. The best scores are boldfaced.}
	\resizebox{\textwidth}{!}{%
		\renewcommand{\arraystretch}{1.0}
		\begin{tabular}{c@{\hspace{11pt}}c@{\hspace{11pt}}c|c@{\hspace{9pt}}c@{\hspace{9pt}}c@{\hspace{9pt}}c|c@{\hspace{9pt}}c@{\hspace{9pt}}c@{\hspace{9pt}}c|c@{\hspace{9pt}}c@{\hspace{9pt}}c@{\hspace{9pt}}c|c@{\hspace{9pt}}c@{\hspace{9pt}}c@{\hspace{9pt}}c}
			\toprule
			\multicolumn{3}{c|}{\multirow{1}{*}{\centering Methods}} & \multicolumn{4}{c|}{CUHK-SYSU}  & \multicolumn{4}{c|}{PRW}& \multicolumn{4}{c|}{MovieNet-PS}& \multicolumn{4}{c}{Average}               \\
			\cmidrule(r){1-19}
			
			\multirow{2}{*}{\centering $\mathcal{L}_{\text{dkd}}$}    & \multirow{2}{*}{\centering $\mathcal{L}_{\text{rkd}}^+$}     & \multirow{2}{*}{\centering $\mathcal{L}_{\text{rim}}$}  & \multicolumn{2}{c|}{Detection}  & \multicolumn{2}{c|}{Re-ID} & \multicolumn{2}{c|}{Detection}  & \multicolumn{2}{c|}{Re-ID} & \multicolumn{2}{c|}{Detection}  & \multicolumn{2}{c|}{Re-ID} & \multicolumn{2}{c|}{Detection}  & \multicolumn{2}{c}{Re-ID} \\
			\cline{4-19}
			\multicolumn{3}{c|}{} & Recall & AP & mAP&Top-1& Recall & AP & mAP&Top-1& Recall & AP & mAP&Top-1& Recall & AP & mAP&Top-1\\
			\midrule
			\centering  & \centering        & \centering & 55.0 & 52.3 & 73.1 & {75.1} & 61.1 & {60.0} & 11.7 & 57.2 & \textbf{91.4} & \textbf{87.1} & \textbf{40.1} & \textbf{82.3} & 69.2 & 66.5 & 41.6 & 71.5 \\
			
			\centering \cmark     & \centering        & \centering & 79.5 & 77.9 & 85.1 & {86.8} & 91.9 & {89.6} & 20.7 & 74.9 & 82.3 & 76.1 & 34.1 & 78.7 & 84.6 & 81.2 & 46.6 & 80.1                    \\  
			
			\centering\cmark           & \centering  \cmark & \centering & 81.2 &79.4 & 90.6  &  91.9 & 92.9 & 90.3 & 39.4 & 82.6 & 85.0 & 78.1 & 34.8 & 79.0 & 86.4 & 82.6 & 55.0 & 84.5  \\
			
			\centering\cmark           & \centering   & \centering \cmark & 81.0 & 79.2 & 89.4 & 90.9 & 92.8 &
			90.4 & 36.5 & 80.6 & 83.7 & 76.7 & 33.5 & 78.7  & 85.8 & 82.1 & 53.1 & 83.4  \\
			
			\centering\cmark          & \centering\cmark  & \centering\cmark & \textbf{81.4} & \textbf{79.7}  & \textbf{91.2} & \textbf{92.6} & \textbf{93.9} & {\textbf{91.3}} & \textbf{42.2}  & \textbf{83.2} & 85.1 & 78.4  & 34.1 & 78.6 & \textbf{86.8} & \textbf{83.1} & \textbf{55.8}  & \textbf{84.8} \\
			\bottomrule
		\end{tabular}%
	}
	\label{tab:ablation_study}
\end{table*}

\subsection{Ablation Study}

\subsubsection{Losses}
In Table~\ref{tab:ablation_study}, we first evaluate the effect of the three losses, the detection knowledge distillation loss $\mathcal{L}_{\text{dkd}}$, the re-ID knowledge distillation loss $\mathcal{L}_{\text{rkd}}^+$, and the rehearsal-based instance matching loss $\mathcal{L}_{\text{rim}}$, respectively. 
Note that detaching all the losses is the same as the FineTune method since the proposed losses are designed to work only when the old data are available.
We see that adding each loss improves the performance, respectively. Specifically, when we use $\mathcal{L}_{\text{dkd}}$, the detection performance is largely increased from that of the FineTune method on the old datasets, and the re-ID performance is also increased accordingly. 
On the other hand, $\mathcal{L}_{\text{rkd}}^+$ and $\mathcal{L}_{\text{rim}}$ slightly increase the detection performance of using $\mathcal{L}_{\text{dkd}}$, but significantly improve the re-ID performance on the old datasets of CUHK-SYSU and PRW by huge margins, demonstrating the effectiveness to preserve the re-ID knowledge in the old domains.

\begin{table}[t]
    \centering
    \caption{Performance comparison when using the transformer based method of COAT~\cite{yu2022cascade} as the backbone network.}
    \renewcommand{\arraystretch}{1}
    \resizebox{0.484\textwidth}{!}{%
    \begin{tabular}{c|p{0.9cm}p{0.9cm}|p{0.9cm}p{0.9cm}|p{0.9cm}p{0.9cm}}
      \toprule
      \multirow{2}{*}{\centering Methods} & \multicolumn{2}{c|}{CUHK-SYSU} & \multicolumn{2}{c|}{PRW} & \multicolumn{2}{c}{MovieNet-PS} \\
      \cline{2-7}
      & \centering mAP & \centering Top-1 & \centering mAP & Top-1 & \centering mAP & Top-1  \\
      \midrule
      FineTune     & \centering  68.5  & \centering 71.3  & \centering 11.6  & 54.8 & \centering \textbf{38.2}  & \textbf{82.4}  \\ 
      Proposed & \centering \textbf{92.4} & \centering \textbf{91.1} & \centering \textbf{40.7} & \textbf{83.0} & \centering 36.0 & 78.7 \\
      \bottomrule 
    \end{tabular}%
    }
    \label{tab:coat_result}
\end{table}

\begin{table}[t]
    \centering
    \caption{Effect of using the prototype features in the old LUT for re-ID knowledge distillation.}
    \renewcommand{\arraystretch}{1}
    \resizebox{0.484\textwidth}{!}{%
    \begin{tabular}{c|p{0.9cm}p{0.9cm}|p{0.9cm}p{0.9cm}|p{0.9cm}p{0.9cm}}
      \toprule
      \multirow{2}{*}{\centering Methods} & \multicolumn{2}{c|}{CUHK-SYSU} & \multicolumn{2}{c|}{PRW} & \multicolumn{2}{c}{MovieNet-PS} \\
      \cline{2-7}
      & \centering mAP & \centering Top-1 & \centering mAP & Top-1 & \centering mAP & Top-1  \\
      \midrule
      Intra-batch     & \centering 90.2   & \centering 91.7  & \centering 40.9  & 82.3 & \centering 33.3  & 78.5 \\ 
      Old prototype & \centering \textbf{91.2} & \centering \textbf{92.6} & \centering \textbf{42.2} & \textbf{83.2} & \centering \textbf{34.1} & \textbf{78.6} \\
      \bottomrule 
    \end{tabular}%
    }
    \label{tab:for_ablation}
\end{table}

\begin{table}[t]
    \caption{Effect of the exemplar data sampling schemes.}
	\centering
		\renewcommand{\arraystretch}{1}
		\resizebox{0.482\textwidth}{!}{%
			\begin{tabular}{c|p{0.9cm}p{0.9cm}|p{0.9cm}p{0.9cm}|p{0.9cm}p{0.9cm}}
      \toprule
      \multirow{2}{*}{\centering Methods} & \multicolumn{2}{c|}{CUHK-SYSU} & \multicolumn{2}{c|}{PRW} & \multicolumn{2}{c}{MovieNet-PS} \\
      \cline{2-7}
      & \centering mAP & \centering Top-1 & \centering mAP & Top-1 & \centering mAP & Top-1  \\
      \midrule
      Max BBox   & \centering \textbf{91.6}   & \centering 92.3  &  \centering 39.0   & 83.0 &  \centering 34.0  & \textbf{80.9}  \\
      Max ID     & \centering 91.3  & \centering 92.2  & \centering 39.5   & 82.8  &  \centering  \textbf{34.8} & 80.0  \\
      Random     & \centering 90.6   & \centering  92.0 & \centering 40.8 & 83.2  &  \centering 33.8  & 80.2 \\  
      Uniform & \centering 91.2 & \centering \textbf{92.6} & \centering \textbf{42.2} & \textbf{83.2} & \centering 34.1 & 78.6 \\
      \bottomrule 
			\end{tabular}%
		}
		\label{tab:exemplar_sampling}
\end{table}

\subsubsection{Baseline Network}
It is worth to note that the proposed method can be applied to any baseline network of person search. We conducted the additional experiment by implementing the proposed method on the transformer based architecture of COAT~\cite{yu2022cascade}. Table~\ref{tab:coat_result} shows the results where we see that the proposed method significantly improves the performance compared to that of the FineTune method.

\subsubsection{Old Prototype-Based Knowledge Distillation}
Table~\ref{tab:for_ablation} shows the effect of using the old prototype features for re-ID knowledge distillation. 
The conventional method, Intra-batch, estimates the distributions of the feature similarity with respect to all the detected proposals within a mini-batch. On the other hand, the proposed method matches the distributions of the feature similarity by using the prototype features of all identities stored in the old LUT, and thus provides better results than the Intra-batch scheme.

\subsubsection{Exemplar Data Sampling}
Table~\ref{tab:exemplar_sampling} shows the results of using different sampling schemes to compose the exemplar data from the old datasets.
`Max BBox' samples the images that have the top 2\% largest numbers of ground truth bounding boxes. `Max ID' samples the images that have the top 2\% largest numbers of person instances with identity labels. `Random' samples 2\% images randomly from the old datasets. 
We see that different sampling schemes provide similar performances to one another, and selected the uniform sampling that yields a slightly better performance of the old knowledge preservation compared to the other ones.

\subsection{Limitation}
The proposed method stores 2\% of the old data into the exemplar memory the typical rehearsal (replay) based methodology of lifelong learning~\cite{rebuffi2017icarl,shieh2020continual, wu2021generalising, ge2022lifelong}. Therefore, the size of the exemplar memory increases as we have more and more datasets. As a future research topic, we will investigate other methodologies of lifelong learning that address the limitation of using the exemplar memory. 

\section{Conclusion}
In this paper, we proposed a novel LPS framework where the model needs to be incrementally trained on the new datasets while preserving the knowledge of the old datasets. We implemented the knowledge distillation between the old and new models based on the rehearsal methodology by using the representative prototype features of the labeled foreground persons as well as the hard background proposals in the old exemplar data. We also designed the rehearsal-based instance matching loss to improve the discrimination ability by using the unlabeled person instances in addition to the prototype features. Experimental results evaluated on three datasets of person search showed that the proposed method achieves significantly better performance of lifelong learning compared with the existing methods, and successfully prevents the knowledge forgetting in the old domains. We expect this pioneering work would encourage further research for practical LPS applications.

\vfill

\end{document}